\documentclass[letterpaper, 10 pt, conference]{ieeeconf}  

\IEEEoverridecommandlockouts                              

\usepackage{booktabs}  
\usepackage{array, booktabs}
\usepackage{pdfpages}
\usepackage{graphicx}  
\usepackage{capt-of}  
\usepackage{multirow}
\usepackage{pifont}
\usepackage{todonotes}
\usepackage{ifthen}

\usepackage{xspace}
\usepackage{caption}  
\usepackage{url}
\usepackage{xcolor}
\usepackage{float}
\newcommand{\sys}{{MALMM\xspace}}
\newcommand{\ntask}{{9 }}

\title{\LARGE \bf
\sys{}: 
Multi-Agent Large Language Models\\
for Zero-Shot Robotic Manipulation
}

\author{Harsh Singh$^{*}$$^{\dagger1}$, Rocktim Jyoti Das$^{\dagger1}$, Mingfei Han$^{1}$, Preslav Nakov$^{1}$, Ivan Laptev$^{1}$%
\thanks{*Corresponding author.}%
\thanks{$\dagger$Equal contribution.}%
\thanks{$^{1}$Mohamed bin Zayed University of Artificial Intelligence, UAE}%
\thanks{email: firstname.lastname@mbzuai.ac.ae,}%
\thanks{Project Page: \protect\url{https://malmm1.github.io/}}%
}

\begin{document}
\maketitle
\thispagestyle{empty}
\pagestyle{empty}

\begin{abstract}

Large Language Models (LLMs) have demonstrated remarkable planning abilities across various domains, including robotic manipulation and navigation. 
While recent work in robotics deploys LLMs for high-level and low-level planning, existing methods often face challenges with failure recovery and suffer from hallucinations in long-horizon tasks. 
To address these limitations, we propose a novel multi-agent LLM framework, \textit{M}ulti-\textit{A}gent Large \textit{L}anguage \textit{M}odel for \textit{M}anipulation (\sys{}).
Notably, \sys{} distributes planning across three specialized LLM agents, namely high-level planning agent, low-level control agent, and a supervisor agent. 
Moreover, by incorporating environment observations after each step, our framework effectively handles intermediate failures and enables adaptive re-planning. Unlike existing methods, \sys{} does not rely on pre-trained skill policies or in-context learning examples and generalizes to unseen tasks. 
In our experiments, \sys{} demonstrates excellent performance in solving previously unseen long-horizon manipulation tasks, and outperforms existing zero-shot LLM-based methods in RLBench by a large margin. 
Experiments with the Franka robot arm further validate our approach in real-world settings.

\end{abstract}

\section{INTRODUCTION}
Robotic manipulation has seen impressive advancements, enabling agents to handle increasingly complex tasks with greater precision and efficacy.
Current solutions, however, often struggle with generalization, in particular when using imitation learning for policy training~\cite{goyal2023rvt, gervet2023act3d}. Such methods typically excel at specific tasks but lack the adaptability to handle new tasks. One major drawback of imitation learning is the labor-intensive and time-consuming process for data collection, which limits the scalability of resulting policies. Moreover, training task-specific manipulation policies typically require thousands of training episodes~\cite{goyal2023rvt, gervet2023act3d}, making the approach computationally expensive and inefficient. 
To cope with the generalization, robotics policies should demonstrate a deeper understanding of their environment. This involves recognizing and grounding relevant objects and understanding the relationships between them~\cite{gao2024physically}. Equipped with this knowledge, policies can then plan and execute actions more efficiently while adapting to changes in the environment and new task requirements. 

Recent advancements in LLMs have demonstrated remarkable generalization and reasoning capabilities across diverse domains such as commonsense, mathematical, and symbolic reasoning~\cite{davis2024mathematics}. These models, particularly when scaled to billions of parameters, exhibit emergent abilities to break down complex tasks into simpler steps through structured reasoning techniques such as \textit{chain of thought}~\cite{Wei2022ChainOT}.
LLMs have already shown promise in high-level task planning in different domains, suggesting their potential for flexible and versatile robotic manipulation. 
Yet, recent LLM-based methods for robotic planning
face multiple challenges. One important issue is the tendency of LLMs to produce incorrect high-level plans and low-level control. Additionally, LLMs suffer from hallucinations in long-context generation~\cite{Maharana2024EvaluatingVL}, which is often observed in closed-loop LLM systems. As a result, they may disregard geometric constraints and the parameters of predefined functions or may even lose the sight of the goal. 

\begin{figure}[t!]
  \centering
  \includegraphics[trim={0cm 12.5cm 20cm 0},clip, width=1\linewidth]{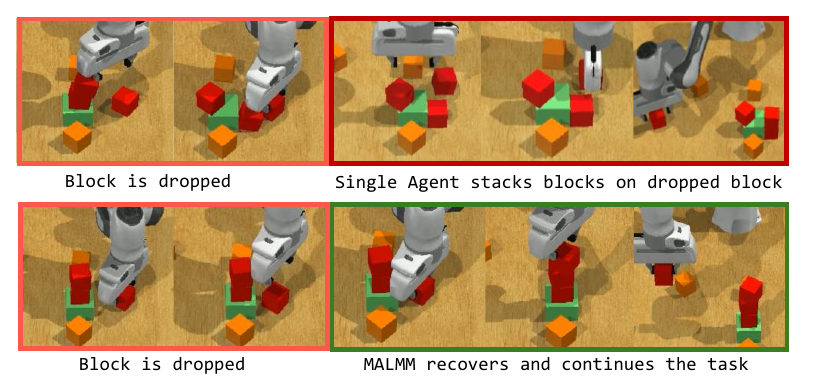}
  \vspace{-.4cm}\\
  \caption{
  Examples of executing \textit{``Stack four blocks at the green target area"} task by the Single Agent LLM (top) and our Multi-Agent MALMM framework (bottom). 
  MALMM recovers after dropping one block and continues stacking above the target area, while the Single Agent mistakenly continues stacking blocks on top of the dropped block.
  }
  \label{fig:qualitative}
  \vspace{-0.3cm}
\end{figure}

\begin{figure*}[t!]
  \centering
  \includegraphics[trim={0cm 5.0cm 9.0cm 0},clip, width=0.90\linewidth]{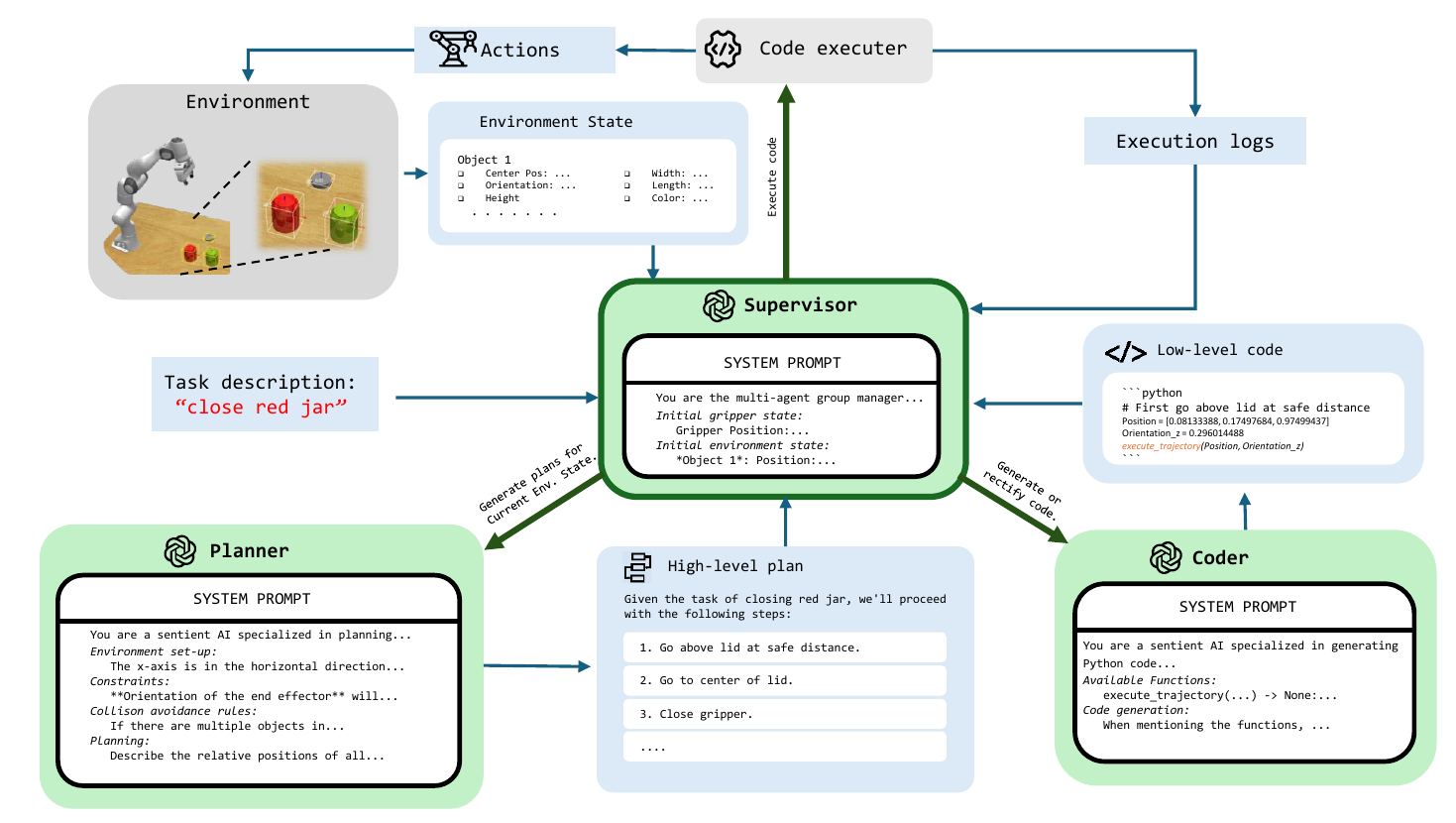}
  \vspace{-0.0cm}\\
  \caption{An overview of our multi-agent system,\textbf{ \sys{}}, which consists of three LLM agents—\textbf{Planner}, \textbf{Coder}, and \textbf{Supervisor}—and a \textbf{Code executor} tool. Each agent operates with a specific system prompt defining its role: (1) the \textbf{Planner} generates high-level plans and replans in case of intermediate failures, (2) the \textbf{Coder} converts these plans into low-level executable code, and (3) the \textbf{Supervisor} coordinates the system by managing the transitions between the Planner, the Coder, and the Code executor.}
  \label{fig:pipeline}
  \vspace{-0.3cm}
\end{figure*}


In this work, we propose a \underline{M}ulti-\underline{A}gent Large \underline{L}anguage \underline{M}odel for \underline{M}anipulation (\sys{}) to leverage the collective intelligence and the specialized skills of multiple agents for complex manipulation tasks. Our framework incorporates agents dedicated to high-level planning, low-level code generation and a supervisor that oversees transitions between other agents and tools.
We show that through the use of multiple specialized agents in a multi-agent setup, we are able to mitigate the hallucination issues observed in the case of a single agent, as shown in Fig.~\ref{fig:qualitative}. 

Our contributions can be summarized as follows: 
\begin{itemize}

\item We introduce the first multi-agent LLM framework for robotic manipulation \sys{}, equipped with specialized agents that bring collaborative and role-specific capabilities to handle unseen, diverse, and complex manipulation tasks.
\item We demonstrate the advantages of the proposed multi-agent framework through systematic ablation studies on tasks with varying horizons and complexity. 
\item We evaluate \sys{} in challenging zero-shot settings both in a simulation and in the real world,  and we show substantial improvements over state-of-the-art methods. 

        

\end{itemize}

\section{RELATED WORK}
\subsection{Language Grounded Robotics}
Language instructions enable the definition of complex robotics tasks with compositional goals~\cite{Thomason2015LearningTI} and support scalable generalization to new tasks \cite{Jang2022BCZZT}. The literature around language grounding is vast, ranging from classical tools such as lexical analysis, formal logic, and graphical models to interpreting language instructions \cite{Thomason2015LearningTI, Kollar2010GroundingVO}. 
Recently, much effort has focused on adopting the impressive 
capabilities of LLMs to language grounding in robotics~\cite{Jang2022BCZZT}.  Additionally, recent advancements have benefited from pre-trained LLMs thanks to their open-world knowledge, tackling more challenging tasks such as 3D robotic manipulation and leveraging code generation capabilities to produce high-level, semantically-rich procedures for robot control.

\subsection{LLM for Robotics}
Language Models have been used for various robotics purposes including the definition of reward functions~\cite{Wang2024RLVLMFRL}, task planning~\cite{Kwon2023LanguageMA, Huang2022InnerME, liu2023llm+}, failure summarization and guiding language-based planners~\cite{Liu2023REFLECTSR}, and policy program synthesis~\cite{Liang2022CodeAP}. VoxPoser~\cite{Huang2023VoxPoserC3} and Language to Rewards~\cite{Yu2023LanguageTR} used LLMs for generating reward regions for assisting external trajectory optimizers in computing trajectory. 
Our work is most related to methods using LLMs for manipulation planning. 
Most of such work~\cite{ Huang2022InnerME, Huang2023VoxPoserC3} relies on pre-trained skills, motion primitives, and trajectory optimizers and has focused primarily on high-level planning. The closest to our approach is \textit{Language Models as Zero-Shot Trajectory Generators}~\cite{Kwon2023LanguageMA}, which deployed a language model to generate high-level plans and then convert these plans into low-level control. However, LLMs suffer from hallucinations, which affect long-horizon task planning. Moreover, \cite{Kwon2023LanguageMA} assumed the correct execution of each step and did not account for occasional failures or unforeseen changes in the environment. Our evaluation shows sizable improvements of \sys{} over~\cite{Kwon2023LanguageMA} thanks to its multiple specialized agents and intermediate environment feedback.
\begin{figure*}[ht!]
  \centering
\includegraphics[trim={0.5cm 5cm 14.5cm 0},clip, width=0.31\linewidth]
{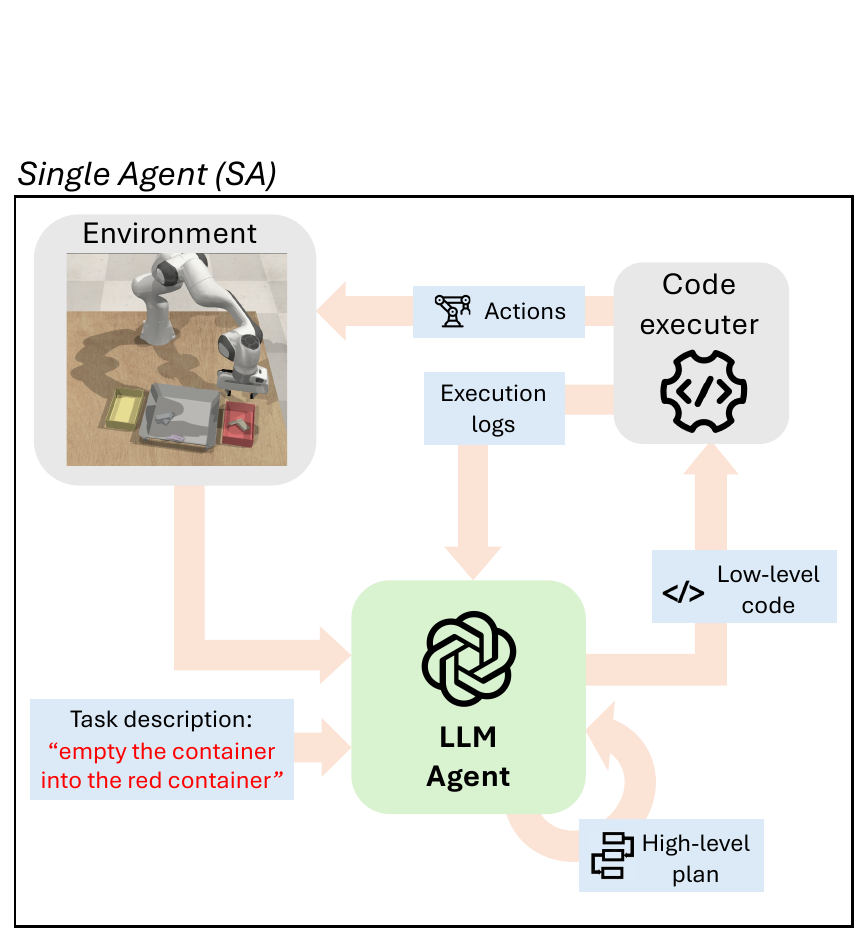}
\includegraphics[trim={0.7cm 5cm 12.8cm 0},clip, width=0.34\linewidth]
{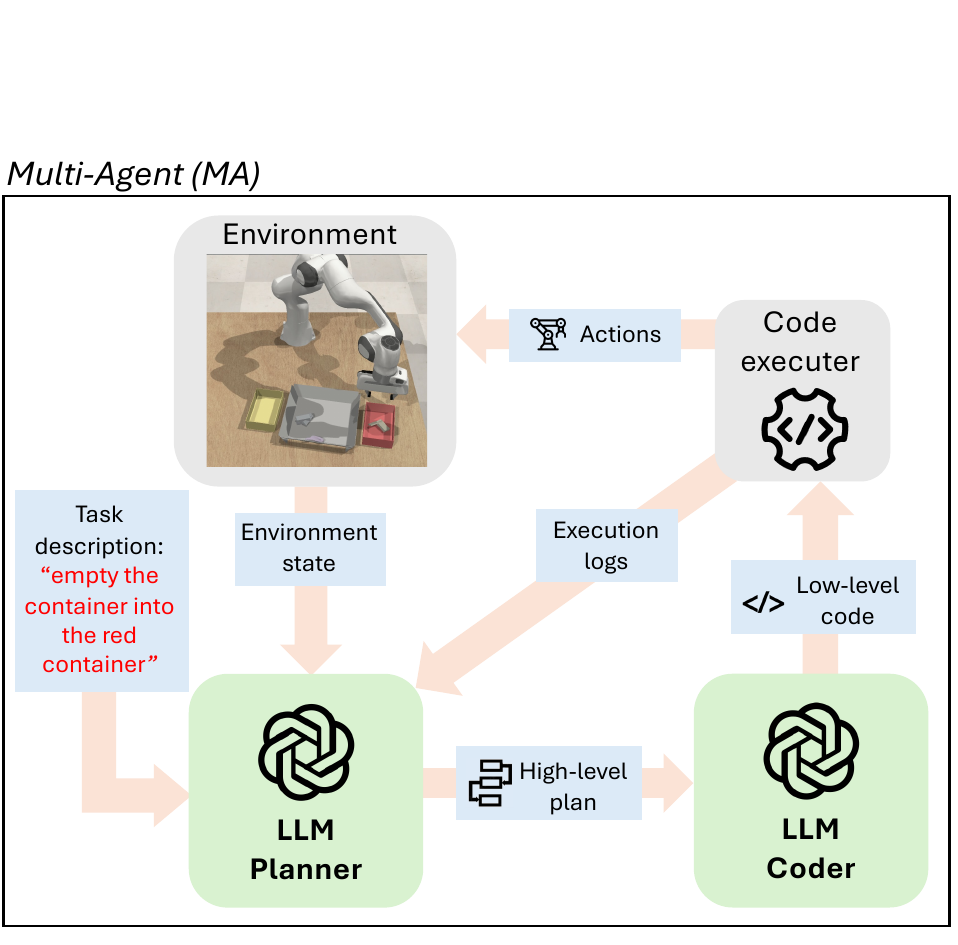}
\includegraphics[trim={0.5cm 4.5cm 12.9cm 0},clip, width=.33\linewidth]
{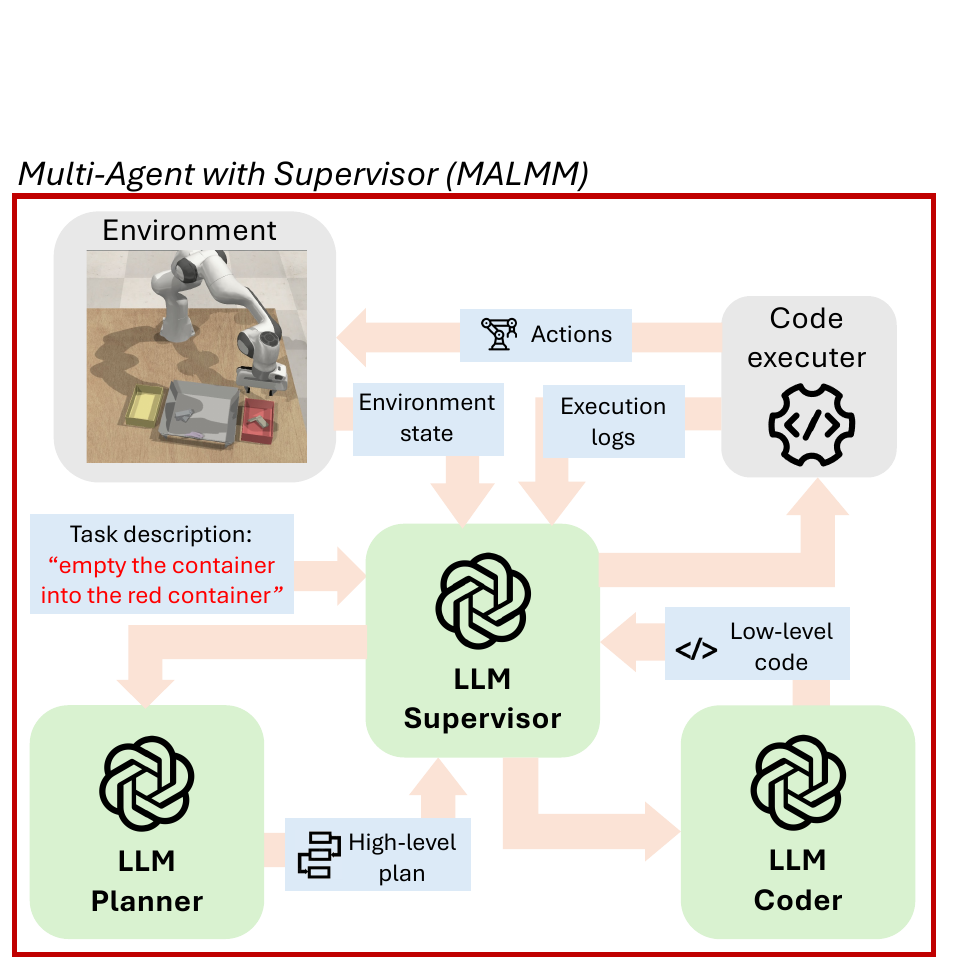}\vspace{-.15cm}\\
(a) \hspace{5.3cm} (b) \hspace{5.3cm} (c)\vspace{-.2cm}\\
  \caption{Agents for robotic manipulation: The figure illustrates three LLM-based manipulation frameworks:  \textbf{SA}, \textbf{MA}, and \textbf{\sys{}}, with the different number of agents in each framework. All three frameworks begin by receiving an input command and the initial environment observation. Each framework iteratively generates a high-level plan along with corresponding low-level code. After each intermediate step, the frameworks use updated environment observation to detect
failures and replan as needed until the task is completed.}
  \label{fig:overview}
  \vspace{-0.4cm}
\end{figure*}

\subsection{LLM-Based Multi-Agents in Robotics}
Recently, several studies have focused on using LLMs for sequential decision-making and complex reasoning tasks~\cite{Bubeck2023SparksOA}. There is also an emerging field of using multiple agents driven by LLMs to solve complex tasks, including robotic manipulation, in a collaborative manner~\cite{Mandi2023RoCoDM}. In most of the work in robotics, LLMs have been used in multi-robot collaboration and communication. Moreover, Reinforcement Learning (RL) policy agents are also used in collaboration with LLMs, which  
limits their generalization to new tasks and environments \cite{Dasgupta2023CollaboratingWL}. In contrast, we propose a multi-agent framework, MALMM, which incorporates three role-specific LLM agents and enables zero-shot execution of previously unseen robotic manipulation tasks.

\section{METHOD}

LLMs have recently emerged as a universal tool for a variety of tasks, including robotic navigation and manipulation~\cite{Huang2022VisualLM, Huang2023VoxPoserC3, Liang2022CodeAP, Kwon2023LanguageMA}. Such models exhibit surprising generalization abilities and support zero-shot solutions for new tasks. While several recent methods have explored LLMs for high-level task planning~\cite{ Huang2022LanguageMA, Huang2022InnerME}, other methods attempted to bring the power of LLMs to the low-level trajectory planning~\cite{Liang2022CodeAP, Kwon2023LanguageMA}. We follow these works and leverage LLMs to precisely control the end-effector for solving complex and previously unseen manipulation tasks, without relying on any predefined skill library for sub-tasks in the manipulation process.

Trajectory planning is a complex problem that requires reasoning about the precise shape, position, and semantics of the objects as well as understanding object interactions. To solve this problem, LLM agents can be instructed to handle information about objects and robots by {\em text prompts}. To harness the complexity of the problem, here we propose to address the manipulation problem with multiple LLM agents. Inspired by recent work on multi-agent LLMs, we design agents specialized in different aspects of the manipulation problem and connect them into a framework. Below, we describe our proposed 
\textbf{MALMM} framework in detail (see Fig.~\ref{fig:pipeline}).



   
\subsection{Preliminaries}

Our goal is to design a LLM framework capable of solving previously unseen manipulation tasks defined by natural language instructions.
We assume access to a robotic environment supporting manipulation actions of closing the gripper, opening the gripper, and changing the gripper's position and orientation around the z axis. Environment observations are obtained as described in Sec.~\ref{sec:observations}. 

\subsection{\sys{}: Multi-Agent Manipulation}
\label{sec:multi-agent}
The core motivation of our work is to investigate how a multi-agent framework can leverage the collaborative and role-specific reasoning capabilities of multiple LLM agents to complete complex manipulation tasks.
Below, we first introduce a single-agent architecture and then we propose its multi-agent extensions.
The single and multi-agent architectures considered in our work are illustrated in Fig.~\ref{fig:overview}.
\smallskip

\noindent\textbf{Single Agent (SA).}
 SA is adopted from \cite{Kwon2023LanguageMA} by environment-specific (RLBench \cite{James2019RLBenchTR} or real-world) prompt tuning. Prompting LLMs to interpret natural language instructions in the form of executable primitive actions has been shown successful for a variety of tasks, including image understanding~\cite{suris2023vipergpt} and numerical reasoning~\cite{chen2022program}.
Similarly, LLMs can be prompted to interpret embodied tasks, e.g.,~{\em open a wine bottle} and convert them into a sequence of primitive manipulation actions. 
A version of such a system with a single LLM agent is outlined in Fig.~\ref{fig:overview}(a). 
Here, an LLM is first prompted to break down the language instruction into a sequence of steps. It then uses its code generation capabilities to translate each step into executable Python code, using predefined functions to control the end-effector. This code is then sent to a Python interpreter that executes the steps in the environment. After each step, the LLM receives new observations from the environment and proceeds in a loop with planning and code generation until meeting termination criteria.
\smallskip


\noindent\textbf{Multi-Agent (MA).}
A Single Agent performing multiple roles struggles to excel in all of them. To address this issue, we propose two specialized LLM agents with shorter role-specific contexts: the \emph{Planner} and the \emph{Coder}, see Fig.~\ref{fig:overview}(b). The \emph{Planner} breaks down the language instructions into a sequence of manipulation steps while the \emph{Coder} iteratively translates these steps into an executable Python code. 
After each intermediate step, the \emph{Planner} detects potential failures and re-plans according to new observations of the environment.
\smallskip


\noindent\textbf{Multi-Agent with a Supervisor.}
%
Our final multi-agent architecture, \sys{} extends MA with a \emph{Supervisor} agent that coordinates the \emph{Planner}, the \emph{Coder}, and the \emph{Code Executor}, as shown in Fig.~\ref{fig:overview}(c).
The \emph{Supervisor} decides which agent or tool to activate next based on the input instructions, the roles of the individual agents, the environment observations, and the entire chat history of the active agents.

\subsection{Multi-Agent Prompting}
Each agent is provided with a task-agnostic system prompt. 
The agents rely solely on their internal world knowledge for reasoning and decision-making. For prompt construction, we draw inspiration from~\cite{Kwon2023LanguageMA} and its study of LLM-based trajectory generation for robotic manipulation. We adapt the prompt according to our environments (RLBench and real-world). Note that unlike other recent work~\cite{Liang2022CodeAP, Huang2023VoxPoserC3}, we do not provide the agents with any examples for in-context learning and apply \sys{} to new tasks without any changes, i.e.,~in a zero-shot mode.

Each agent's prompt is specifically designed to suit its role. Since LLMs require step-by-step reasoning to solve tasks, the \emph{Planner} 
is prompted to generate steps that define the intermediate goals needed to complete the task. \sys{} perceives the environment observations directly from the simulator or by analyzing the RGBD (see Sec.~\ref{sec:observations}). Therefore, the \emph{Planner} is given a detailed description of the environment's coordinate system, enabling it to interpret the directions from the gripper's perspective. Given the limited exposure of LLMs to grounded physical interaction data in their training, LLM agents often fail to account for potential collisions. To address this, we include generic collision-avoidance rules in the \emph{Planner} prompts. Moreover, to handle intermediate failures, primarily due to collisions or missed grasps, we prompt the \emph{Planner} to evaluate action based on the previous and current environment observations after each intermediate step, and to replan if necessary. The prompt used for the Planner agent is shown in Figure~\ref{fig:planner_prompt}.




The prompt for the Coder, as shown in Figure~\ref{fig:coder_prompt},
includes information about the expected input and output for all available functions—\textit{execute\_trajectory()}, \textit{open\_gripper()}, \textit{close\_gripper()}, and \textit{check\_task\_completion()}—as well as guidelines to avoid syntactic and semantic errors, which are common in code generated by LLMs. 
Finally, the Supervisor agent, as presented in Figure~\ref{fig:supervisor_prompt},
is prompted to manage the workflow, coordinating the transitions between the LLM agents, the Planner the Coder, and the Code Executor to ensure successful task completion.

\subsection{Environment Observations}
\label{sec:observations}
LLMs trained on textual inputs cannot directly perceive or interpret 3D environments. Our agents receive information about the environment either from the internal state space of a simulator or from RGBD.

\smallskip\noindent
\textbf{State-space observations.} In this setup, the LLM agents have direct access to the simulator's state information. The observations are provided as 3D bounding boxes (object dimensions, center position, and orientation), along with object colors and the gripper’s position, orientation, and open/closed state. The execution logs for \textit{empty container} task presented in  Appendix\ref{app:output} show input observation format as CURRENT ENVIRONMENT STATE for \sys{}. The same format is also used for Single Agent and Multi-Agent presented in Section~\ref{sec:multi-agent}. 

\begin{figure*}[ht]
  \centering
  \includegraphics[trim={0 10.2cm 0 0},clip, width=1\linewidth]
  {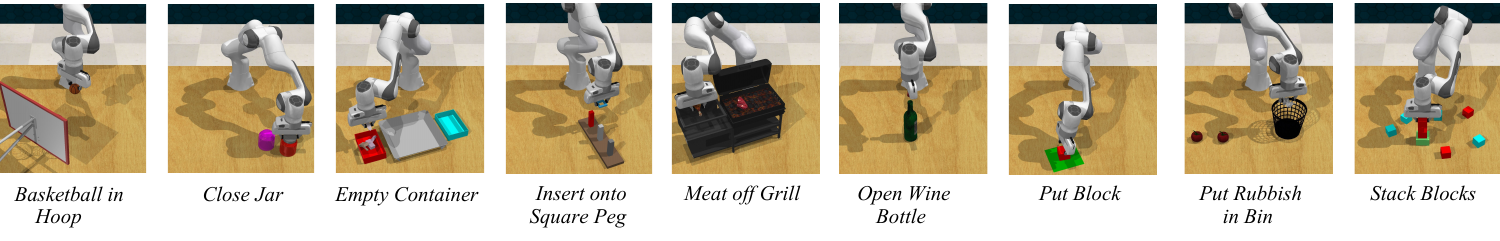}
  \vspace{-0.7cm}\\
  \caption{Illustration of the \textbf{nine} RLBench~\cite{James2019RLBenchTR} tasks used in our evaluation, featuring diverse tasks with varying task horizons and different object shapes.}
  \label{fig:tasks}
  \vspace{-0.2cm}
\end{figure*}

\smallskip\noindent
\textbf{Visual observations.} To apply \sys{} in real-world settings, we restrict observations to the front-facing RGB-D sensor (as illustrated in Fig.~\ref{fig:tasks}), from which we obtain both RGB images and 3D point clouds. We then use pre-trained foundation models to extract information about scene objects.
To this end, we employ \textit{gpt-4-turbo} to derive a list of objects relevant to the instruction text and RGB image, and subsequently utilize LangSAM~\cite{medeiros2024langsegmentanything} to produce segmentation masks for these objects, such as \emph{block} or \emph{red jar}.
We then segment the 3D object point clouds by projecting the 2D segmentation masks into the 3D space. To compute accurate object-centric grasping poses, we apply the M2T2~\cite{Yuan2023M2T2MM} model and predict grasps given 3D point clouds of source objects.
We use the obtained gripper poses to control the gripper during grasping. 
To facilitate object placement, we estimate the 3D bounding box of the target object in the environment. For example, in the task \emph{close the red jar}, the target object would be the \emph{red jar}. We extract the 3D bounding box directly from the object's point cloud, and use it to guide the placement process. We leverage these visual observations while conducting experiments in both simulated and real-world settings. Please refer to Appendix\ref{app:vis_state} for details on obtaining visual observations.


\section{EXPERIMENTS}

\begin{table*}[h]
\caption{Success rate for zero-shot evaluation on RLBench~\cite{James2019RLBenchTR}: The table highlights the best-performing method for each task in \textbf{bold} and the second-best-performing method is \underline{underlined}.  \small{\textbf{Symbol:} † denotes LLaMA-3.3-70B as a base model. All other methods use GPT-4-Turbo as a base model.} }
\vspace{-0.1cm}
\centering
\scalebox{1.1}{%
\begin{tabular}{lcccccccccc}
    \toprule
    \multirow{2}{*}{\textbf{Methods}} & \textbf{Basketball} & \textbf{Close} & \textbf{Empty} & \textbf{Insert} & \textbf{Meat} & \textbf{Open} & \textbf{Put} & \textbf{Rubbish} & \textbf{Stack} & \multirow{2}{*}{\textbf{Avg}}  \\ 
    & \textbf{in Hoop} & \textbf{Jar} & \textbf{Container} & \textbf{in Peg} & \textbf{off Grill} & \textbf{Bottle} & \textbf{Block} & \textbf{in Bin} & \textbf{Blocks} &  \\ \midrule  
    CAP~\cite{Liang2022CodeAP} & 0.00 & 0.00 & 0.00 & 0.08 & 0.00 & 0.00 & 0.76 & 0.00 & 0.00 & 0.09  \\
    VoxPoser~\cite{Huang2023VoxPoserC3} & 0.20 & 0.00 & 0.00 & 0.00 & 0.00 & 0.00 & 0.36 & \underline{0.64} & \underline{0.32} & 0.17\\
    Single Agent (SA)~\cite{Kwon2023LanguageMA} & \underline{0.52} & \underline{0.40} & \underline{0.36} & \underline{0.24} & \underline{0.44} & \underline{0.80} & \underline{0.92} & 0.48 & 0.20 & \underline{0.50} \\
    \sys{} (†) & 0.84 & \textbf{0.88} & 0.60 & \textbf{0.80} & 0.64 & 0.84 & 0.84 & 0.56 & 0.32 & 0.70\\
     \sys{} & \textbf{0.88} & 0.84 & \textbf{0.64} & 0.68 & \textbf{0.92} & \textbf{0.96} & \textbf{1.00} & \textbf{0.80} & \textbf{0.56} & \textbf{0.81}\\
    \bottomrule
\end{tabular}
}
\label{tab:model-comp}
\end{table*}

We evaluate the accuracy of \sys{} in a zero-shot settings, i.e.~when solving diverse set of previously unseen tasks defined by a short text description.

\vspace{-1mm}
\subsection{Implementation Details.}

We use \textit{gpt-4-turbo}\footnote{The experiments were conducted in June 2024 using the gpt-4-turbo-2024-04-09 snapshot.}~\cite{openai2023gpt4} to drive the LLM agents in all our experiments. Additionally, we report the results of \sys{} using LLaMA-3.3-70B~\cite{Dubey2024TheL3} to demonstrate the performance of our framework with an open-source LLM model.
For developing the multi-agent framework, we used AutoGen \cite{Wu2023AutoGenEN}, which is an open-source programming library for building AI agents and facilitates collaboration between multiple agents to solve complex tasks. To perform zero-shot evaluation, we do not fine-tune our agents, and we use no training data for in-context learning. We initially developed our prompts for the \textit{Stack Blocks} task and used them for other tasks without any task-specific tuning. \sys{} generates 3D waypoints, while the trajectories are computed and executed using a motion planner, following the approach commonly used in RLBench.
Our code, prompts, and additional results are available from the project webpage~\cite{malmmwebpage}.

\vspace{-2mm}
\subsection{Environment and Tasks.}
We conduct the simulation in CoppeliaSim, interfaced via PyRep, using Franka Panda robot with a parallel gripper. The setup incorporates RLBench~\cite{James2019RLBenchTR}, a robot learning benchmark that provides various language-conditioned tasks with specified success criteria. For evaluation, we sampled \ntask RLBench tasks with 25 samples per task, featuring diverse object poses, object shapes, and task horizons. Fig.~\ref{fig:tasks} shows snapshots of the nine considered tasks.

 
In the real-world setup, we evaluated five tasks on a tabletop using a 7-DOF Franka Emika Panda Research 3 robot equipped with a parallel jaw gripper. Three of these tasks are identical to those that use vision-based observations from the simulator, while two are new. We use an Intel RealSense D435i RGB-D camera to capture the frontal view and the panda-py~\cite{elsner2023taming} library to control the robot arm.


\subsection{Baselines.} We compare our approach to three state-of-the-art zero-shot LLM-based manipulation methods: Code as Policies (CAP)~\cite{Liang2022CodeAP},  VoxPoser~\cite{Huang2023VoxPoserC3}, and SA~\cite{Kwon2023LanguageMA}. For CAP\footnote{\url{https://code-as-policies.github.io/}}, we adapt its official implementation for use with RLBench. We used the official implementation of VoxPoser\footnote{\url{https://voxposer.github.io/}} without any modifications. Our SA baseline is a version of ``Language Models as Zero-Shot Trajectory Generators''~\cite{Kwon2023LanguageMA} with environment-specific (RLBench or real-world) prompt
tuning. Additionally, for fair comparison between the SA and \sys{}, the prompts we provided to both of them were equivalent. The only difference is that in the case of \sys{}, the information and the instructions in the Single-Agent prompt are distributed between the \emph{Supervisor}, the \emph{Planner}, and the \emph{Coder} according to their respective roles. All the baselines use \textit{gpt-4-turbo} as an LLM. The prompts for Single Agent and \sys{} are provided in Appendix\ref{app:single_prompts} and Appendix\ref{app:malmm_prompts} respectively.
\subsection{Results for Zero-Shot Evaluation}

Table \ref{tab:model-comp} presents the results for the three baseline methods, along with our proposed \sys{}, across \ntask different tasks. From the table, we observe that \sys{} outperforms all baselines across all \ntask tasks, including long-horizon tasks such as \emph{stack blocks} and \emph{empty container}, as well as tasks involving complex shapes, such as \emph{meat off grill} and \emph{rubbish in bin}. Moreover, the Code as Policies is able to generate a successful trajectory for only two tasks. This limited success is because the original Code as Policies implementation relied on few-shot examples to perform well on tasks involving regularly shaped objects. In our evaluation, we replaced these few-shot examples with coordinate definitions and detailed instructions about the functions available for the LLM to call. However, Code as Policies completely failed in this zero-shot setting, which was also reported in \cite{Kwon2023LanguageMA}.

VoxPoser~\cite{Huang2023VoxPoserC3}, which generates 3D voxel maps for value functions to predict waypoints, successfully generated trajectories for three tasks with good accuracy. For two of these tasks, it was the second-best performing method. However, its performance did not generalize well to other tasks. 

Both the SA and our proposed multi-agent framework, \sys{}, successfully generated trajectories for all \ntask tasks. However, \sys{} consistently outperformed the SA approach by using different agents for specific roles, enabling it to generate accurate high-level plans and low-level code while mitigating hallucinations. In addition to the \textit{gpt-4-turbo} experiments, we evaluated MALMM using the open-source LLaMA-3.3-70B~\cite{Dubey2024TheL3}; the results are presented in Table~\ref{tab:model-comp}. Although there is a drop in performance compared to \textit{gpt-4-turbo}, MALMM, with LLaMA-3.3-70B, outperforms the existing baselines by a sizable margin. We provide qualitative comparison between Single Agent and \sys{} for the \textit{Stack Blocks} task in Appendix\ref{app:qual_results}.

\subsection{Multi-Agent Ablation}

We performed ablation for each of the components (agents) in \sys{} in order to evaluate how each of them contributes to the overall performance. 
We considered two tasks, namely {\em stack blocks} and {\em empty container}, and report the results in Table~\ref{tab:multi-agent-ablation}.


We first analyzed the importance of the intermediate environment feedback. To this end, we considered the Single Agent~(SA) setting and removed the environment feedback provided after each intermediate step. In this setup, LLM generates the full manipulation plan at once and executes it without revisions. 
As shown in Table~\ref{tab:multi-agent-ablation}, the SA without environment feedback exhibits 12\% and 24\% drop in performance for the \emph{stack blocks} and \emph{empty container} tasks, respectively. By analyzing the failure cases of both methods, we observed that the environment feedback provided after each intermediate step crucially affected the agent's ability to detect unforeseen situations and recover from failures such as collisions and inaccurate grasping.

\begin{table}[t]
\caption{Ablation study assessing the impact of different components in \sys{}: environment feedback, \emph{Planner} (\textbf{P}), \emph{Coder} (\textbf{C}), and \emph{Supervisor} (\textbf{S}).}
\setlength{\tabcolsep}{1.5mm} 
\renewcommand{\arraystretch}{0.9} 
\footnotesize 
\begin{tabular}{lcccccc}
\toprule
\multirow{2}{*}{\bf{Agents}} & \bf{Environment}  & \multirow{2}{*}{\bf{P}}  &\multirow{2}{*}{\bf{C}}  &\multirow{2}{*}{\bf{S}}  & \bf{Stack} & \bf{Empty} \\ 
& \bf{Feedback} &&&&\bf{Blocks} & \bf{Container} \\ \midrule
Single Agent (SA) & \ding{55} & \ding{55} & \ding{55} & \ding{55} & 0.08 & 0.12 \\ 
Single Agent (SA) & \ding{51} & \ding{55} & \ding{55} & \ding{55} & 0.20 & 0.36 \\ 
Multi-Agent (MA)  & \ding{51} & \ding{51} & \ding{51} & \ding{55} & 0.36 & 0.48 \\ 
\sys{}          & \ding{51} & \ding{51} & \ding{51} & \ding{51} & 0.56 & 0.64 \\ 
\bottomrule
\vspace{-0.7cm}\\
\end{tabular}
\label{tab:multi-agent-ablation}
\end{table}

We next validated the advantage of the Multi-Agent architecture with separate LLM agents for planning and code generation. As shown in Table~\ref{tab:multi-agent-ablation}, the MA, consisting of a dedicated \emph{Planner} and \emph{Coder}, demonstrated 16\% and 12\% performance improvement over SA for the two tasks respectively. This can be attributed to the inherent limitations of LLMs in managing very long context conversations~\cite{Maharana2024EvaluatingVL}.
In the SA setup, where a single LLM is responsible for both the high-level planning and the low-level code generation, the agent must handle an extensive context, particularly for long-horizon tasks. This often leads to errors such as failing to account for collisions with other objects, omitting the input arguments for the predefined functions, using variables before they were initialized, and even forgetting the specified goal. In contrast, the MA system mitigates these issues by dividing the workload among specialized agents. The \emph{Planner} and the \emph{Coder} agents in the MA setting focus on specific roles through specialized prompts and communicate with each other, thus reducing the likelihood of errors and hallucinations, in particular for longer tasks. 


Our initial Multi-Agent system pre-defines the cyclic sequence of the \emph{Planner}, the \emph{Coder}, and the \emph{Coder Executor}, see Fig.~\ref{fig:overview}(b), assuming that each agent correctly completes its task. However, hallucinations may occur even within multi-agent systems~\cite{Banerjee2024LLMsWA}. For example, the \emph{Coder} may miss the variable initialization resulting in compilation errors or incomplete sub-goal code generation, such as producing code only for approaching the object without grasping it. 
In such situations, the \emph{Coder} may need to be re-executed in order to correct possible errors before passing the control to the \emph{Planner}. To automate this process, we introduced a \emph{Supervisor} agent that dynamically re-routes the execution process to the next agent based on the input instruction, the entire communication history of all active agents, and the role descriptions of all agents rather than following a fixed sequence. This adaptive approach is at the core of our MALMM framework, and it improves the performance of the dual-agent MA setup by 20\% in the \textit{'stack blocks'} task and 16\% in the \textit{'empty container'} task, respectively, as shown in Table~\ref{tab:multi-agent-ablation}.
\subsection{\sys{} is Better at Long-Horizon Planning}
\begin{figure}[ht]
  \centering
  \includegraphics[width=1\linewidth]{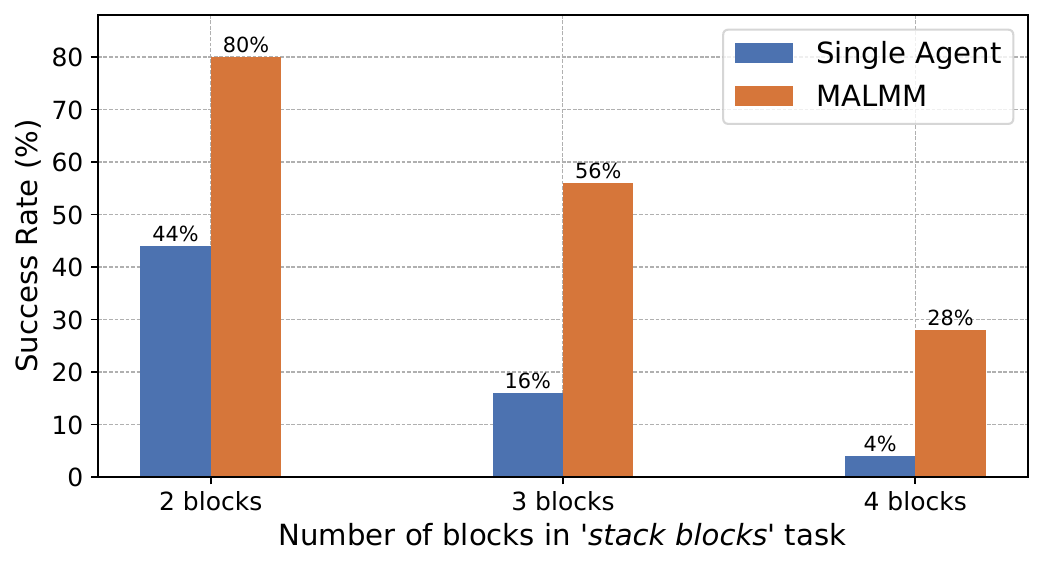}
  \vspace{-0.3cm}
  \caption{Comparison of \textbf{Single Agent} vs. \textbf{\sys{}} for variations of the \textit{stack blocks} task that require stacking 2, 3, or 4 blocks on top of each other.}
  \label{fig:mas_vs_sa}
  \vspace{-0.3cm}
\end{figure}

To validate the effectiveness of \sys{} in long-horizon tasks, we created three variations of the \textit{'stack blocks'} task, each with a different number of blocks, and compared the performance of \sys{} to the Single Agent setup. The results in Fig.~\ref{fig:mas_vs_sa} indicate that while the Single Agent setup struggles with stacking 3 and 4 blocks, \sys{} substantially outperforms SA, in particular for tasks that require longer planning. Fig.~\ref{fig:qualitative} illustrates an example of failure recovery with key steps by MALMM compared to SA on the stack block task.

\begin{figure*}[ht]
  \centering
  \includegraphics[trim={0 9.5cm 0 0},clip, width=1\linewidth]{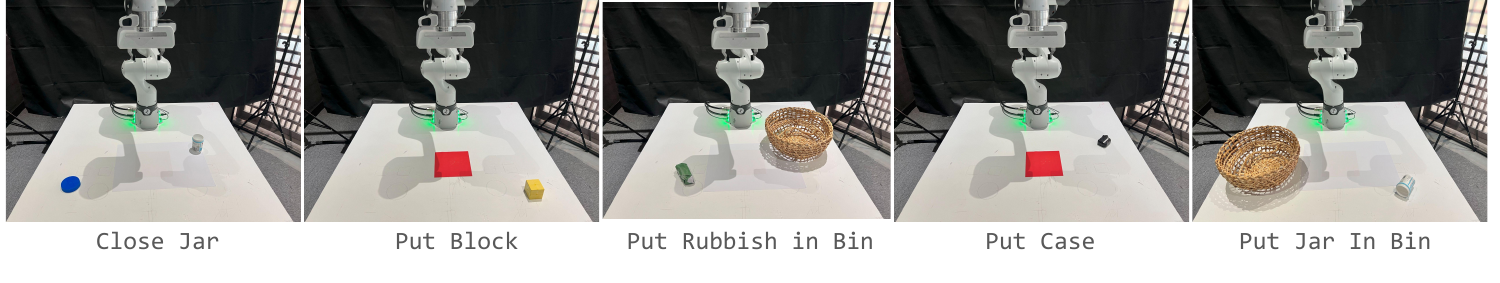}
  \vspace{-0.7cm}\\
  \caption{Illustration of the \textbf{five} real-world tasks used in our evaluation.}
  \label{fig:tasks_real}
  \vspace{-0.7cm}
\end{figure*}

\begin{figure*}[ht]
  \centering
  \includegraphics[trim={0.2cm 15cm 0.2cm 0cm},width=1\linewidth]{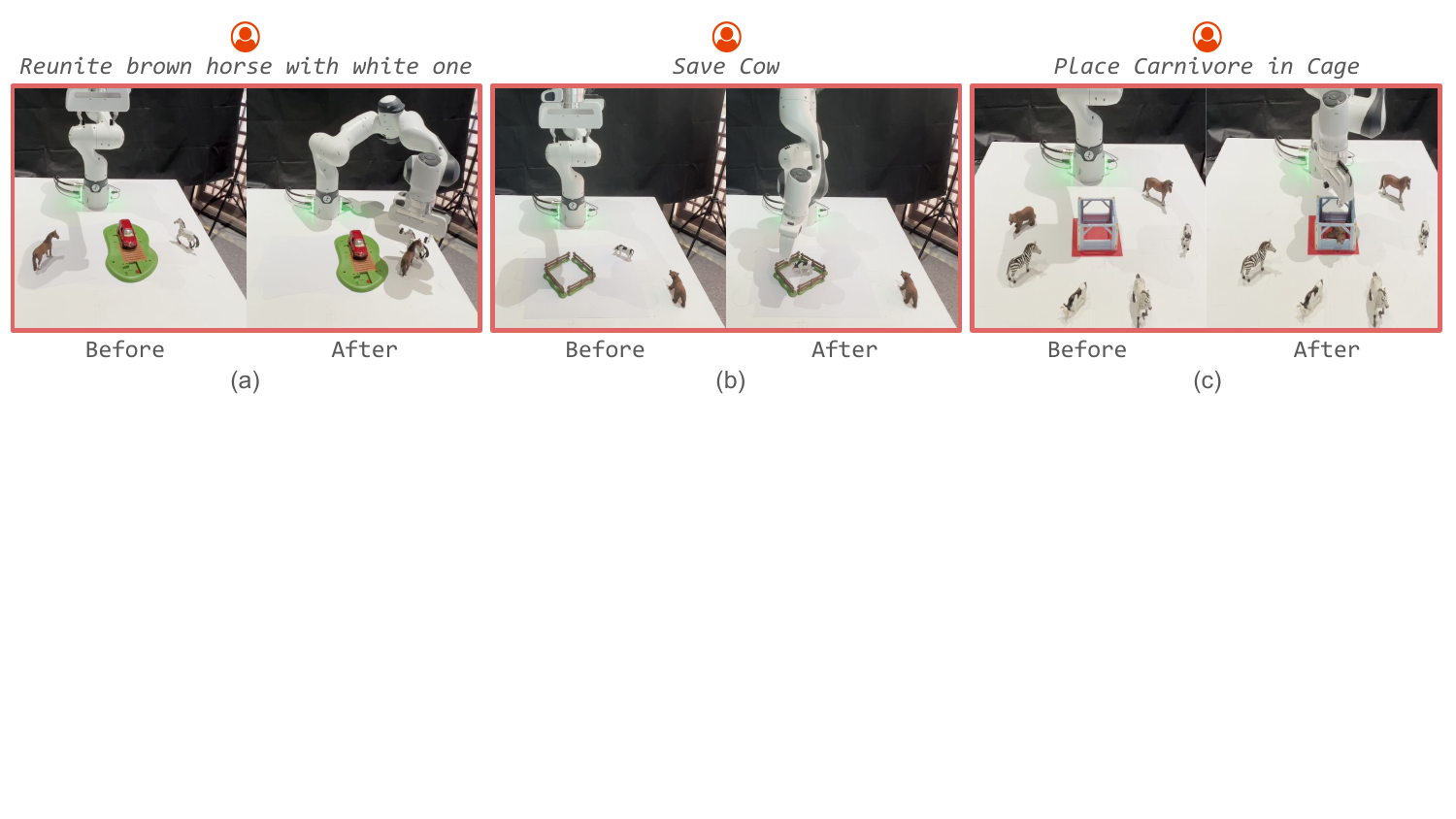}
  \vspace{5cm}
  \caption{MALMM performs zero-shot manipulation on three unseen tasks in a real world, each guided by \textit{high-level} user instructions. (a) “\textit{Reunite the brown horse with the white one.}” The environment contains a brown horse, a white horse, a road, and a car; the goal is to place the brown horse near the white horse. (b) “\textit{Save Cow.}” The environment contains a bear, a cow, and a fenced enclosure; the goal is to place the cow inside the enclosure. (c) “\textit{Place Carnivore in Cage.}” The environment contains two horses, two cows, one zebra, a bear, and a cage; the goal is to place the bear inside the cage. In each scenario, the left image shows the initial arrangement and the right image shows the final arrangement after MALMM completes the instructed task.}
  \label{fig:qual_real}
  \vspace{-0.2cm}
\end{figure*}

\subsection{Results for Vision-Based Observations} 

In order to approach real-world settings where direct access to the environment states is not available, we next perform experiments in simulation restricted to vision-based observations in the form of 3D point clouds. 
We evaluate the performance of \sys{} and compare it to the Single Agent setup on three tasks: (\emph{ii})~\emph{put block}, (\emph{ii})~\emph{rubbish in bin}, and (\emph{iii})~\emph{close jar}. 
Consistently with our previous state-based experiments, the results in \sys{} in Table~\ref{tab:vision} show sizable improvements over the Single-Agent baseline across all three tasks and confirm the advantage of our proposed Multi-Agent framework.

By comparing the results in Tables~\ref{tab:model-comp} and \ref{tab:vision} we observe degradation in performance when switching to vision-based observations. This can be attributed to inaccuracies of the vision-based estimators such as 3D bounding box detection and grasp estimation. Note that our current vision pipeline makes use of single-view scene observations.
A parallel work, Manipulate Anything~\cite{Duan2024ManipulateAnythingAR} showed that a relatively straightforward extension to multi-view settings can reduce the impact of occlusions and yield higher accuracy.

\begin{table}[ht]
\setlength{\tabcolsep}{4mm} 
\centering
\caption{Comparison of the Single Agent and \sys{} in a simulated environment with vision-based observations.}
\begin{tabular}{lccc}
\toprule
\multirow{2}{*}{\bf{Agents}} & \bf{Close} & \bf{Put} & \bf{Rubbish} \\ 
                           & \bf{Jar}   & \bf{Block} & \bf{in Bin} \\ 
\midrule
Single Agent~\cite{Kwon2023LanguageMA} & 0.24 & 0.68 & 0.40 \\ 
\sys{}      & \textbf{0.56} & \textbf{0.84} & \textbf{0.52} \\ 
\bottomrule
\end{tabular}
\vspace{-0.2cm}
\label{tab:vision}
\end{table}

\subsection{Results for the Real-World Experiments} For the real-world robotics setup, we evaluated five tasks as shown in Fig. \ref{fig:tasks_real}: \emph{close jar} (put lid on top of jar),   \emph{put block} (place a block in the red target area), \emph{put rubbish in bin} (place rubbish in the bin), \emph{put case} (place an earbuds case in the red target area),  and \emph{put jar in bin} (place a jar in the bin) -- in ten different initial states, each with both \sys{} and the Single Agent. As shown in Table~\ref{tab:real_world}, consistently with our simulation results, \sys{} outperforms the Single Agent on all five tasks by a sizable margin. It achieved a 40\% higher success rate for both \emph{put case} and \emph{rubbish in bin}, 30\% higher success rate for \emph{put block}, and 20\% higher success rate for \emph{close jar} and \emph{jar in bin}. 
To further demonstrate zero-shot capabilities of our method, Fig. \ref{fig:qual_real} demonstrates successful performance of \sys{} for three new tasks, each defined by \emph{high-level} user instruction.


\begin{table}[h]
\setlength{\tabcolsep}{2mm} 
\centering
\footnotesize 
\caption{Comparison of Single Agent and \sys{} in a real-world Franka robot arm environment.}
\resizebox{\columnwidth}{!}{%
\begin{tabular}{lccccc}
\toprule
\multirow{2}{*}{\bf{Agents}} & \bf{Close} & \bf{Put} & \bf{Rubbish} & \bf{Put} & \bf{Jar}\\ 
                            & \bf{Jar}   & \bf{Block} & \bf{in Bin} & \bf{Case} & \bf{in Bin}\\ 
\midrule
Single Agent~\cite{Kwon2023LanguageMA} & 2/10 & 3/10 & 2/10 & 3/10 & 3/10 \\ 
\sys{} & \textbf{4/10} & \textbf{6/10} & \textbf{6/10} & \textbf{7/10} & \textbf{5/10}\\ 
\bottomrule
\end{tabular}%
}
\vspace{-0.2cm}
\label{tab:real_world}
\end{table}


\section{DISCUSSION}

\subsection{Limitations} Despite its string advantages over SA, \sys{} has several limitations. First, \sys{} relies on three \textit{gpt-4-turbo} agents, making it costly to operate. Using open-source LLMs is possible at the cost of reduced accuracy (cf.~Table~\ref{tab:model-comp}). Second, like other LLM-based planners, \sys{} depends on manual prompt engineering, which impacts its performance. However, advancements in prompting~\cite{Wei2022ChainOT} can reduce these efforts. Finally, \sys{} requires accurate bounding box estimation to determine the correct grasp positions, but as the complexity of the objects increases, the 3D bounding boxes alone may not provide enough information for precise grasping. Our experiments with a vision pipeline in simulation and in a real-world scenario suggest that using pretrained grasping and placement models, such as M2T2~\cite{Yuan2023M2T2MM}, could improve the performance for complex manipulation tasks. Even though \sys{} performs over 6$\times$ better than the SA on the long-horizon task (stacking 4 blocks), it achieves only a 28\% success rate, revealing limitations in handling rich object interactions. Additionally, the vision-based pipeline is not robust to severe occlusions due to its single-view limitation.

\subsection{Conclusion and Future Work}
We explored the use of LLM agents for solving previously unseen manipulation tasks. 
In particular, we proposed the first multi-agent LLM framework for robotics manipulation \sys{} and demonstrated its advantages over single-agent baselines.
Our method uses task-agnostic prompts and requires no in-context learning examples for solving new tasks.
Extensive evaluations, both in simulation and real-world settings, demonstrated excellent results for \sys{} for a variety of manipulation tasks. 
%
%
In future work, we will explore richer object interactions and more complex tasks, including articulated objects. We also aim to use vision-language foundation models to incorporate contextual scene details beyond object bounding boxes.

\bibliographystyle{IEEEtran} %
\bibliography{refs}

\begin{thebibliography}{10}
\providecommand{\url}[1]{#1}
\csname url@rmstyle\endcsname
\providecommand{\newblock}{\relax}
\providecommand{\bibinfo}[2]{#2}
\providecommand\BIBentrySTDinterwordspacing{\spaceskip=0pt\relax}
\providecommand\BIBentryALTinterwordstretchfactor{4}
\providecommand\BIBentryALTinterwordspacing{\spaceskip=\fontdimen2\font plus
\BIBentryALTinterwordstretchfactor\fontdimen3\font minus \fontdimen4\font\relax}
\providecommand\BIBforeignlanguage[2]{{%
\expandafter\ifx\csname l@#1\endcsname\relax
\typeout{** WARNING: IEEEtran.bst: No hyphenation pattern has been}%
\typeout{** loaded for the language `#1'. Using the pattern for}%
\typeout{** the default language instead.}%
\else
\language=\csname l@#1\endcsname
\fi
#2}}

\bibitem{goyal2023rvt}
A.~Goyal, J.~Xu, Y.~Guo, V.~Blukis, Y.-W. Chao, and D.~Fox, ``Rvt: Robotic view transformer for 3d object manipulation,'' \emph{arXiv preprint arXiv:2306.14896}, 2023.

\bibitem{gervet2023act3d}
T.~Gervet, Z.~Xian, N.~Gkanatsios, and K.~Fragkiadaki, ``Act3d: 3d feature field transformers for multi-task robotic manipulation,'' in \emph{Conference on Robot Learning}.\hskip 1em plus 0.5em minus 0.4em\relax PMLR, 2023, pp. 3949--3965.

\bibitem{gao2024physically}
J.~Gao, B.~Sarkar, F.~Xia, T.~Xiao, J.~Wu, B.~Ichter, A.~Majumdar, and D.~Sadigh, ``Physically grounded vision-language models for robotic manipulation,'' in \emph{2024 IEEE International Conference on Robotics and Automation (ICRA)}.\hskip 1em plus 0.5em minus 0.4em\relax IEEE, 2024, pp. 12\,462--12\,469.

\bibitem{davis2024mathematics}
E.~Davis, ``Mathematics, word problems, common sense, and artificial intelligence,'' \emph{Bulletin of the American Mathematical Society}, vol.~61, no.~2, pp. 287--303, 2024.

\bibitem{Wei2022ChainOT}
\BIBentryALTinterwordspacing
J.~Wei, X.~Wang, D.~Schuurmans, M.~Bosma, E.~H. hsin Chi, F.~Xia, Q.~Le, and D.~Zhou, ``Chain of thought prompting elicits reasoning in large language models,'' \emph{ArXiv}, vol. abs/2201.11903, 2022. [Online]. Available: \url{https://api.semanticscholar.org/CorpusID:246411621}
\BIBentrySTDinterwordspacing

\bibitem{Maharana2024EvaluatingVL}
\BIBentryALTinterwordspacing
A.~Maharana, D.-H. Lee, S.~Tulyakov, M.~Bansal, F.~Barbieri, and Y.~Fang, ``Evaluating very long-term conversational memory of llm agents,'' \emph{ArXiv}, vol. abs/2402.17753, 2024. [Online]. Available: \url{https://api.semanticscholar.org/CorpusID:268041615}
\BIBentrySTDinterwordspacing

\bibitem{Thomason2015LearningTI}
\BIBentryALTinterwordspacing
J.~Thomason, S.~Zhang, R.~J. Mooney, and P.~Stone, ``Learning to interpret natural language commands through human-robot dialog,'' in \emph{International Joint Conference on Artificial Intelligence}, 2015. [Online]. Available: \url{https://api.semanticscholar.org/CorpusID:6745034}
\BIBentrySTDinterwordspacing

\bibitem{Jang2022BCZZT}
\BIBentryALTinterwordspacing
E.~Jang, A.~Irpan, M.~Khansari, D.~Kappler, F.~Ebert, C.~Lynch, S.~Levine, and C.~Finn, ``Bc-z: Zero-shot task generalization with robotic imitation learning,'' \emph{ArXiv}, vol. abs/2202.02005, 2022. [Online]. Available: \url{https://api.semanticscholar.org/CorpusID:237257594}
\BIBentrySTDinterwordspacing

\bibitem{Kollar2010GroundingVO}
\BIBentryALTinterwordspacing
T.~Kollar, S.~Tellex, D.~K. Roy, and N.~Roy, ``Grounding verbs of motion in natural language commands to robots,'' in \emph{International Symposium on Experimental Robotics}, 2010. [Online]. Available: \url{https://api.semanticscholar.org/CorpusID:5821491}
\BIBentrySTDinterwordspacing

\bibitem{Wang2024RLVLMFRL}
\BIBentryALTinterwordspacing
Y.~Wang, Z.~Sun, J.~Zhang, Z.~Xian, E.~Biyik, D.~Held, and Z.~M. Erickson, ``Rl-vlm-f: Reinforcement learning from vision language foundation model feedback,'' \emph{ArXiv}, vol. abs/2402.03681, 2024. [Online]. Available: \url{https://api.semanticscholar.org/CorpusID:267499679}
\BIBentrySTDinterwordspacing

\bibitem{Kwon2023LanguageMA}
\BIBentryALTinterwordspacing
T.~Kwon, N.~D. Palo, and E.~Johns, ``Language models as zero-shot trajectory generators,'' \emph{IEEE Robotics and Automation Letters}, vol.~9, pp. 6728--6735, 2023. [Online]. Available: \url{https://api.semanticscholar.org/CorpusID:264289016}
\BIBentrySTDinterwordspacing

\bibitem{Huang2022InnerME}
\BIBentryALTinterwordspacing
W.~Huang, F.~Xia, T.~Xiao, H.~Chan, J.~Liang, P.~R. Florence, A.~Zeng, J.~Tompson, I.~Mordatch, Y.~Chebotar, P.~Sermanet, N.~Brown, T.~Jackson, L.~Luu, S.~Levine, K.~Hausman, and B.~Ichter, ``Inner monologue: Embodied reasoning through planning with language models,'' in \emph{Conference on Robot Learning}, 2022. [Online]. Available: \url{https://api.semanticscholar.org/CorpusID:250451569}
\BIBentrySTDinterwordspacing

\bibitem{liu2023llm+}
B.~Liu, Y.~Jiang, X.~Zhang, Q.~Liu, S.~Zhang, J.~Biswas, and P.~Stone, ``Llm+ p: Empowering large language models with optimal planning proficiency,'' \emph{arXiv preprint arXiv:2304.11477}, 2023.

\bibitem{Liu2023REFLECTSR}
\BIBentryALTinterwordspacing
Z.~Liu, A.~Bahety, and S.~Song, ``Reflect: Summarizing robot experiences for failure explanation and correction,'' \emph{ArXiv}, vol. abs/2306.15724, 2023. [Online]. Available: \url{https://api.semanticscholar.org/CorpusID:259274760}
\BIBentrySTDinterwordspacing

\bibitem{Liang2022CodeAP}
\BIBentryALTinterwordspacing
J.~Liang, W.~Huang, F.~Xia, P.~Xu, K.~Hausman, B.~Ichter, P.~R. Florence, and A.~Zeng, ``Code as policies: Language model programs for embodied control,'' \emph{2023 IEEE International Conference on Robotics and Automation (ICRA)}, pp. 9493--9500, 2022. [Online]. Available: \url{https://api.semanticscholar.org/CorpusID:252355542}
\BIBentrySTDinterwordspacing

\bibitem{Huang2023VoxPoserC3}
\BIBentryALTinterwordspacing
W.~Huang, C.~Wang, R.~Zhang, Y.~Li, J.~Wu, and L.~Fei-Fei, ``Voxposer: Composable 3d value maps for robotic manipulation with language models,'' \emph{ArXiv}, vol. abs/2307.05973, 2023. [Online]. Available: \url{https://api.semanticscholar.org/CorpusID:259837330}
\BIBentrySTDinterwordspacing

\bibitem{Yu2023LanguageTR}
\BIBentryALTinterwordspacing
W.~Yu, N.~Gileadi, C.~Fu, S.~Kirmani, K.-H. Lee, M.~G. Arenas, H.-T.~L. Chiang, T.~Erez, L.~Hasenclever, J.~Humplik, B.~Ichter, T.~Xiao, P.~Xu, A.~Zeng, T.~Zhang, N.~M.~O. Heess, D.~Sadigh, J.~Tan, Y.~Tassa, and F.~Xia, ``Language to rewards for robotic skill synthesis,'' \emph{ArXiv}, vol. abs/2306.08647, 2023. [Online]. Available: \url{https://api.semanticscholar.org/CorpusID:259164906}
\BIBentrySTDinterwordspacing

\bibitem{Bubeck2023SparksOA}
\BIBentryALTinterwordspacing
S.~Bubeck, V.~Chandrasekaran, R.~Eldan, J.~A. Gehrke, E.~Horvitz, E.~Kamar, P.~Lee, Y.~T. Lee, Y.-F. Li, S.~M. Lundberg, H.~Nori, H.~Palangi, M.~T. Ribeiro, and Y.~Zhang, ``Sparks of artificial general intelligence: Early experiments with gpt-4,'' \emph{ArXiv}, vol. abs/2303.12712, 2023. [Online]. Available: \url{https://api.semanticscholar.org/CorpusID:257663729}
\BIBentrySTDinterwordspacing

\bibitem{Mandi2023RoCoDM}
\BIBentryALTinterwordspacing
Z.~Mandi, S.~Jain, and S.~Song, ``Roco: Dialectic multi-robot collaboration with large language models,'' \emph{2024 IEEE International Conference on Robotics and Automation (ICRA)}, pp. 286--299, 2023. [Online]. Available: \url{https://api.semanticscholar.org/CorpusID:259501567}
\BIBentrySTDinterwordspacing

\bibitem{Dasgupta2023CollaboratingWL}
\BIBentryALTinterwordspacing
I.~Dasgupta, C.~Kaeser-Chen, K.~Marino, A.~Ahuja, S.~Babayan, F.~Hill, and R.~Fergus, ``Collaborating with language models for embodied reasoning,'' \emph{ArXiv}, vol. abs/2302.00763, 2023. [Online]. Available: \url{https://api.semanticscholar.org/CorpusID:253180684}
\BIBentrySTDinterwordspacing

\bibitem{Huang2022VisualLM}
\BIBentryALTinterwordspacing
C.~Huang, O.~Mees, A.~Zeng, and W.~Burgard, ``Visual language maps for robot navigation,'' \emph{2023 IEEE International Conference on Robotics and Automation (ICRA)}, pp. 10\,608--10\,615, 2022. [Online]. Available: \url{https://api.semanticscholar.org/CorpusID:252846548}
\BIBentrySTDinterwordspacing

\bibitem{Huang2022LanguageMA}
\BIBentryALTinterwordspacing
W.~Huang, P.~Abbeel, D.~Pathak, and I.~Mordatch, ``Language models as zero-shot planners: Extracting actionable knowledge for embodied agents,'' \emph{ArXiv}, vol. abs/2201.07207, 2022. [Online]. Available: \url{https://api.semanticscholar.org/CorpusID:246035276}
\BIBentrySTDinterwordspacing

\bibitem{James2019RLBenchTR}
\BIBentryALTinterwordspacing
S.~James, Z.~Ma, D.~R. Arrojo, and A.~J. Davison, ``Rlbench: The robot learning benchmark \& learning environment,'' \emph{IEEE Robotics and Automation Letters}, vol.~5, pp. 3019--3026, 2019. [Online]. Available: \url{https://api.semanticscholar.org/CorpusID:202889132}
\BIBentrySTDinterwordspacing

\bibitem{suris2023vipergpt}
D.~Sur{\'\i}s, S.~Menon, and C.~Vondrick, ``Vipergpt: Visual inference via python execution for reasoning,'' in \emph{Proceedings of the IEEE/CVF International Conference on Computer Vision}, 2023, pp. 11\,888--11\,898.

\bibitem{chen2022program}
W.~Chen, X.~Ma, X.~Wang, and W.~W. Cohen, ``Program of thoughts prompting: Disentangling computation from reasoning for numerical reasoning tasks,'' \emph{arXiv preprint arXiv:2211.12588}, 2022.

\bibitem{medeiros2024langsegmentanything}
L.~Medeiros, ``Language segment-anything,'' \url{https://github.com/luca-medeiros/lang-segment-anything}, 2024, accessed: 2024-09-12.

\bibitem{Yuan2023M2T2MM}
\BIBentryALTinterwordspacing
W.~Yuan, A.~Murali, A.~Mousavian, and D.~Fox, ``M2t2: Multi-task masked transformer for object-centric pick and place,'' in \emph{Conference on Robot Learning}, 2023. [Online]. Available: \url{https://api.semanticscholar.org/CorpusID:263629874}
\BIBentrySTDinterwordspacing

\bibitem{openai2023gpt4}
OpenAI, ``{GPT-4} technical report,'' 2023.

\bibitem{Dubey2024TheL3}
\BIBentryALTinterwordspacing
A.~Dubey, A.~Jauhri, A.~Pandey, A.~Kadian, A.~Al-Dahle, A.~Letman, A.~Mathur, A.~Schelten, and A.~Y. et~al, ``The llama 3 herd of models,'' \emph{ArXiv}, vol. abs/2407.21783, 2024. [Online]. Available: \url{https://api.semanticscholar.org/CorpusID:271571434}
\BIBentrySTDinterwordspacing

\bibitem{Wu2023AutoGenEN}
\BIBentryALTinterwordspacing
Q.~Wu, G.~Bansal, J.~Zhang, Y.~Wu, B.~Li, E.~Zhu, L.~Jiang, X.~Zhang, S.~Zhang, J.~Liu, A.~H. Awadallah, R.~W. White, D.~Burger, and C.~Wang, ``Autogen: Enabling next-gen llm applications via multi-agent conversation,'' 2023. [Online]. Available: \url{https://api.semanticscholar.org/CorpusID:263611068}
\BIBentrySTDinterwordspacing

\bibitem{malmmwebpage}
``{MALMM} project webpage,'' \url{https://malmm1.github.io/}.

\bibitem{elsner2023taming}
\BIBentryALTinterwordspacing
J.~Elsner, ``Taming the panda with python: A powerful duo for seamless robotics programming and integration,'' \emph{SoftwareX}, vol.~24, p. 101532, 2023. [Online]. Available: \url{https://www.sciencedirect.com/science/article/pii/S2352711023002285}
\BIBentrySTDinterwordspacing

\bibitem{Banerjee2024LLMsWA}
\BIBentryALTinterwordspacing
S.~Banerjee, A.~Agarwal, and S.~Singla, ``Llms will always hallucinate, and we need to live with this,'' 2024. [Online]. Available: \url{https://api.semanticscholar.org/CorpusID:272524830}
\BIBentrySTDinterwordspacing

\bibitem{Duan2024ManipulateAnythingAR}
\BIBentryALTinterwordspacing
J.~Duan, W.~Yuan, W.~Pumacay, Y.~R. Wang, K.~Ehsani, D.~Fox, and R.~Krishna, ``Manipulate-anything: Automating real-world robots using vision-language models,'' \emph{ArXiv}, vol. abs/2406.18915, 2024. [Online]. Available: \url{https://api.semanticscholar.org/CorpusID:270764457}
\BIBentrySTDinterwordspacing

\bibitem{ester1996density}
M.~Ester, H.-P. Kriegel, J.~Sander, X.~Xu, \emph{et~al.}, ``A density-based algorithm for discovering clusters in large spatial databases with noise,'' in \emph{kdd}, vol.~96, no.~34, 1996, pp. 226--231.

\end{thebibliography}
\newpage
\onecolumn
\appendices
\section*{\textbf{Appendix}}
This appendix provides additional details for the \sys{} approach. We present complete prompts in Appendix\ref{app:prompts}, qualitative results in Appendix\ref{app:qual_results}, RLBench tasks in Appendix\ref{app:rlbench_tasks}, details on the visual observation pipeline in Appendix\ref{app:vis_state} and an example of a full execution log in Appendix\ref{app:output}.

\subsection{Prompts}
\label{app:prompts}
\subsubsection{Single Agent Prompts}
\label{app:single_prompts}
\begin{figure*}[ht!]
  \centering
  \includegraphics[trim={0cm 1.4cm 0cm 0},clip, width=1\linewidth]{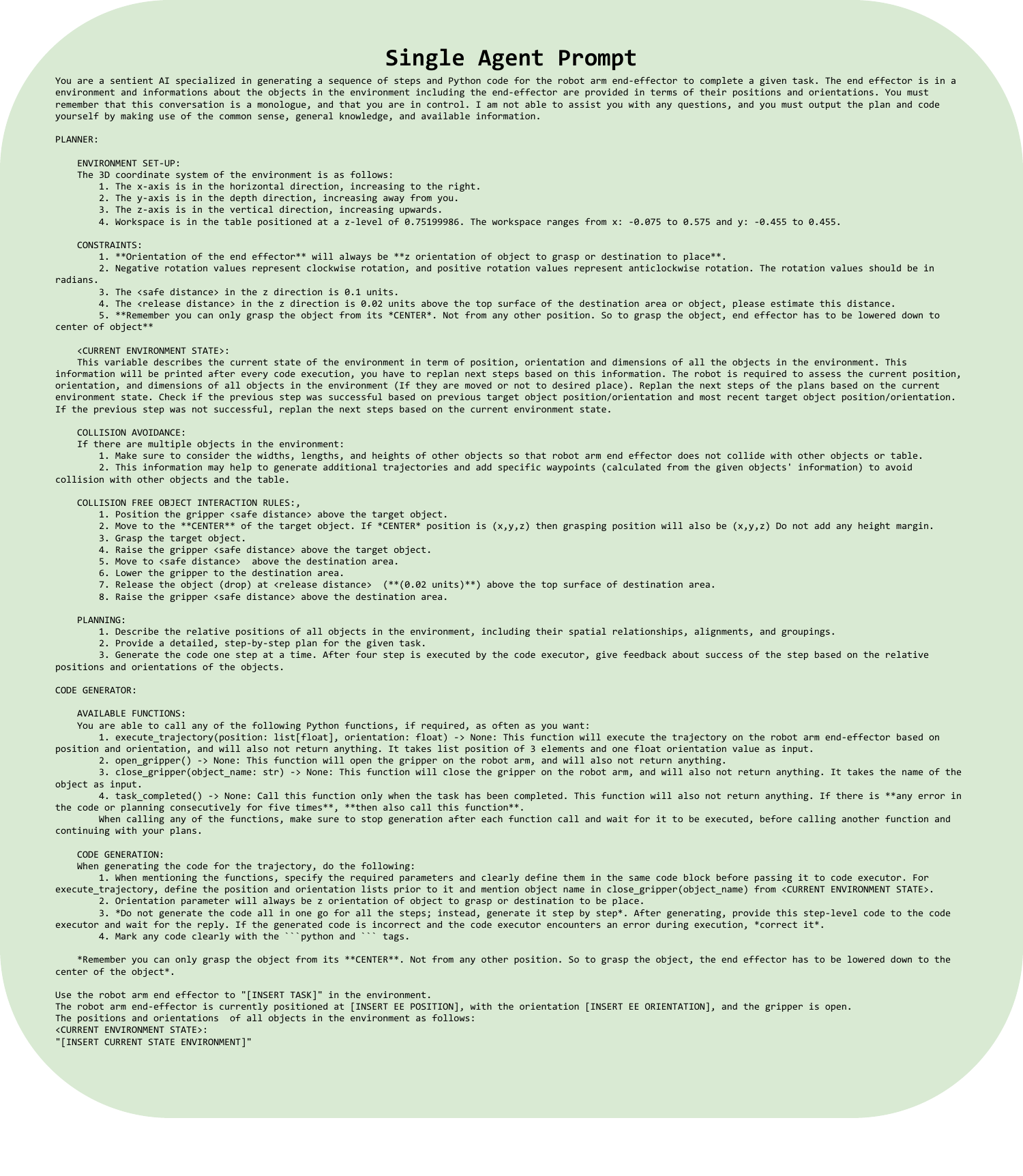}
  \vspace{-1.4cm}\\
  \caption{Prompt for Single Agent.}
  \label{fig:single_agent_prompt}
\end{figure*}

The prompt for the Single Agent baseline is shown in Figure \ref{fig:single_agent_prompt}. There are four variables in this prompt: \textit{[INSERT TASK]} for task instruction, \textit{[INSERT EE POSITION]} for end effector initial position, \textit{[INSERT EE ORIENTATION] }for end effector initial orientation and \textit{[STATE]} for environment state or observation.
\vspace{0.2cm}
\subsubsection{\sys{} Prompts}
 \label{app:malmm_prompts}
\begin{figure*}[ht!]
  \centering
  \includegraphics[trim={0cm 0.7cm 0.0cm 0},clip, width=1\linewidth]{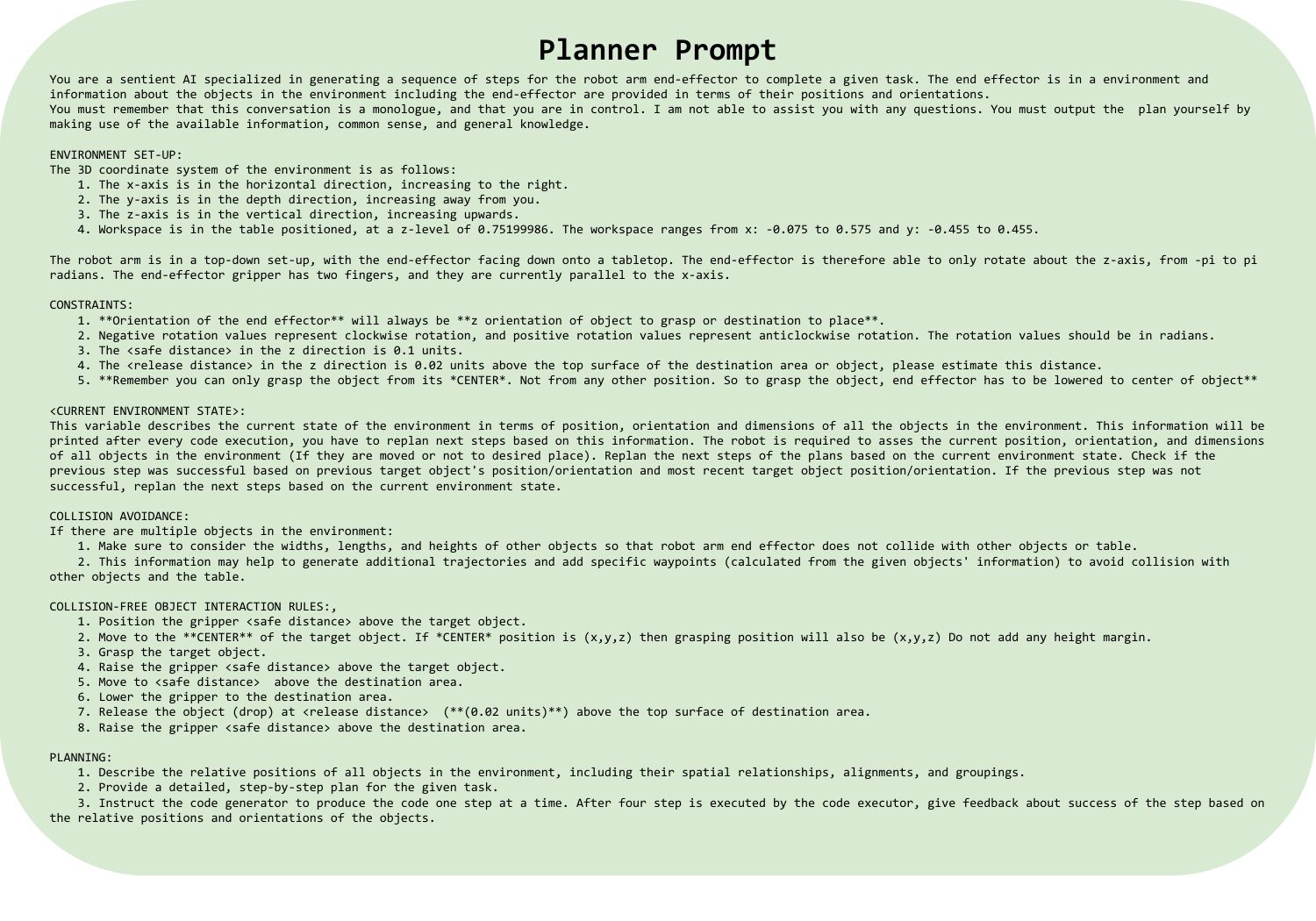}
  \caption{Prompt for Planner of \sys{}}
  \label{fig:planner_prompt}
\end{figure*}

\begin{figure*}[ht!]
  \centering
  \includegraphics[trim={0cm 7.2cm 0.0cm 0},clip, width=1\linewidth]{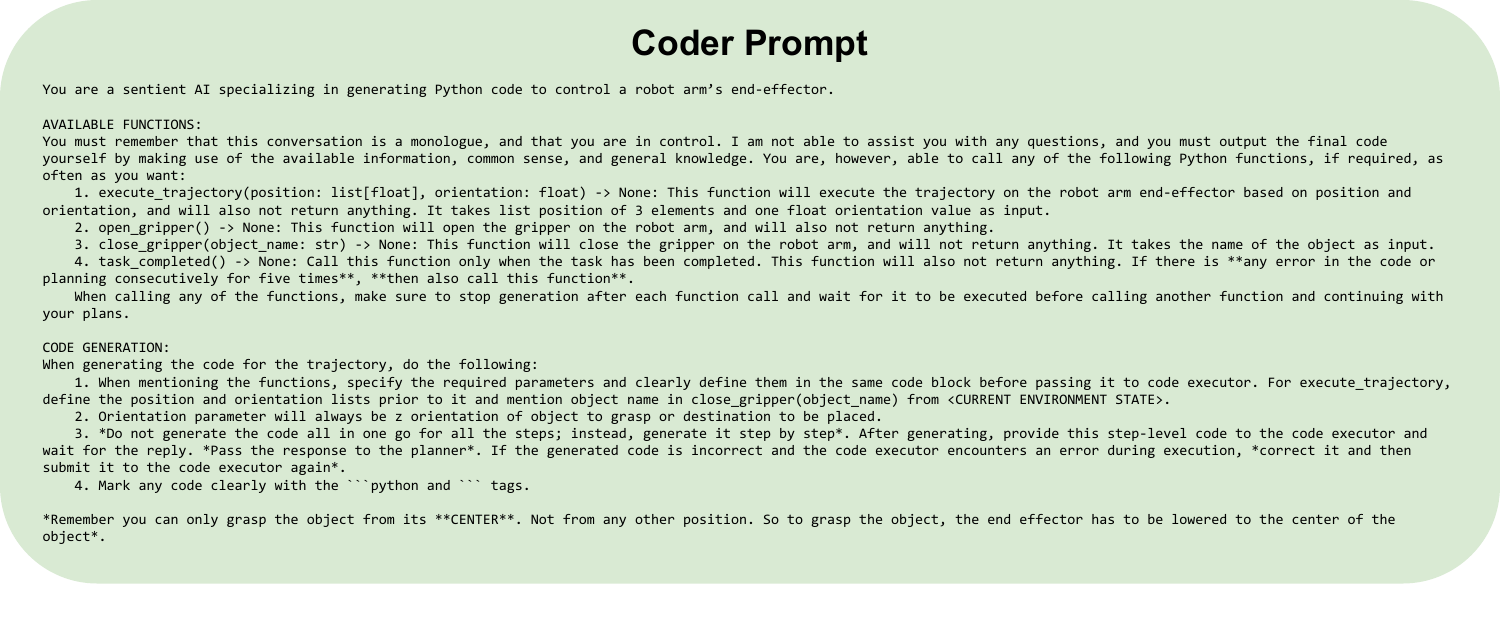}
  \vspace{-0.2cm}\\
  \caption{Prompt for Coder of \sys{}}
  \label{fig:coder_prompt}
\end{figure*}

The Supervisor prompt of \sys{}, shown in Figure \ref{fig:supervisor_prompt}, contains the same four variables as the Single Agent prompt described in Appendix\ref{app:single_prompts}: \textit{[INSERT TASK]}, \textit{[INSERT EE POSITION]}, \textit{[INSERT EE ORIENTATION]}, and \textit{[STATE]}. Additionally, the prompts for Planner and Coder in \sys{} are depicted in Figures \ref{fig:planner_prompt}
 and \ref{fig:coder_prompt} respectively.
\vspace{0.2cm}

\begin{figure*}[ht!]
  \centering
  \includegraphics[trim={0cm 10.5cm 0cm 0},clip, width=1\linewidth]{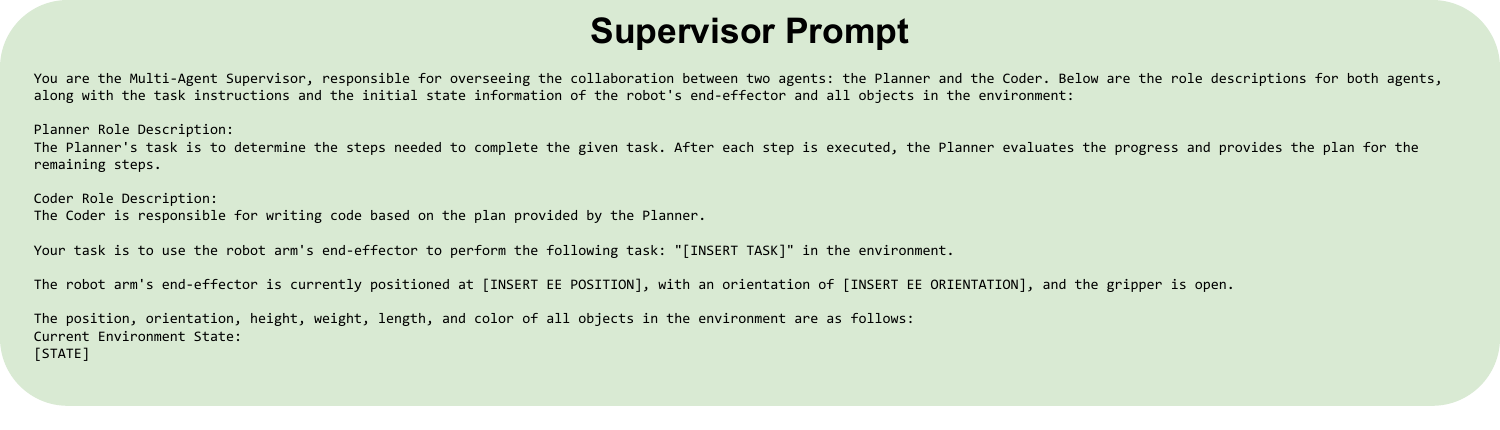}
  \vspace{-0.2cm}\\
  \caption{Prompt for Supervisor of \sys{}}
  \label{fig:supervisor_prompt}
\end{figure*}
\vspace{0.2cm}

\vspace{2cm}
\subsubsection{Multi-Agent Prompts}
 \label{app:ma_prompts}
The Multi-Agent baseline consists of two agents: the Planner and the Coder. The prompts for each agent are the same as those used by \sys{}, as shown in Figure~\ref{fig:planner_prompt} for the Planner and in Figure~\ref{fig:coder_prompt} for the Coder.

\subsection{Single Agent vs \sys{} Qualitative results}
 \label{app:qual_results}
\begin{figure*}[ht!]
  \centering
  \includegraphics[trim={0cm 8cm 0.0cm 0},clip, width=1\linewidth]{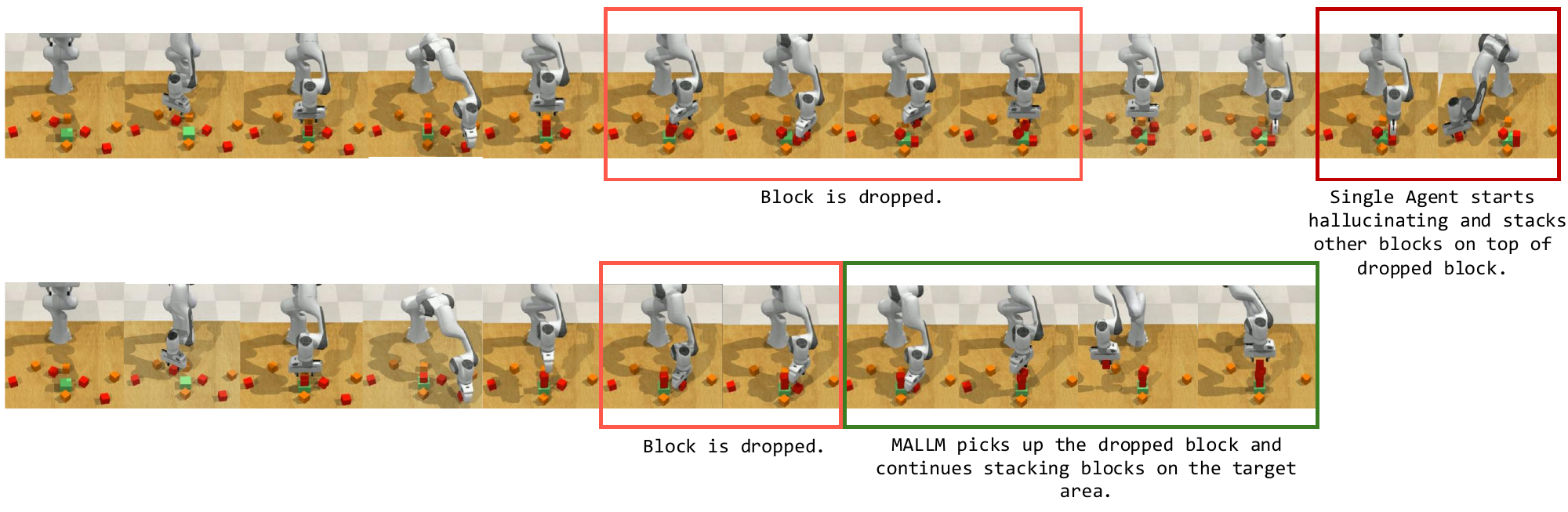}
  \vspace{-0.2cm}\\
  \caption{Qualitative results comparing the Single Agent approach and \sys{} are presented. The figure illustrates examples of the task \textit{"Stack four blocks at the green target area"} performed by the Single Agent LLM (top) and our \sys{} approach (bottom). While both approaches experience failures, \sys{} demonstrates recovery by resuming the stacking process after dropping a block and continuing above the target area. In contrast, the Single Agent mistakenly stacks additional blocks on top of the dropped block.}
  \label{fig:qual_result}
\end{figure*}

Figure~\ref{fig:qual_result} shows qualitative result comparing Single Agent and \sys{}. The sequence of frames on the top row refers to Single Agent, while the bottom row refers to \sys{}. In the case of Single Agent, two blocks were initially stacked correctly but while approaching the third block, the gripper collided with the stacked blocks causing the second block to fall. This subsequently led to hallucinations and ultimately stacking the blocks away from the target area. In contrast, although \sys{} dropped the third block while attempting to place it on top of the stacked blocks on the target area, it was able to successfully detect and rectify the failure, ultimately completing the task by stacking all the blocks on the target area.
\subsection{RLBench Tasks}
\label{app:rlbench_tasks}
We evaluated on nine RLBench tasks which are listed in Table~\ref{tab:tasks} along with the task instruction and success criteria.
\begin{table*}[t]
    \centering
    \caption{Details of the RLBench tasks used for evaluation.}
    \label{tab:tasks}
    \renewcommand{\arraystretch}{1.4} 
    \setlength{\tabcolsep}{10pt} 
    \begin{tabular}{p{0.3\textwidth} p{0.6\textwidth}}
        \toprule
        \textbf{Task Instruction} & \textbf{Details} \\
        \toprule
        
        \textit{Basketball In Hoop} &
        \textit{Task Description:} Put basketball in hoop. \newline
        \textit{Success Criteria:} Basketball passes through hoop. \\
        \midrule

        \textit{Close Jar} &
        \textit{Task Description:} Close the colored jar with a lid. \newline
        \textit{Success Criteria:} Lid is on top of the colored jar. \\
        \midrule

        \textit{Empty Container} &
        \textit{Task Description:} Pick all the objects from the large container and put them into the colored container. \newline
        \textit{Success Criteria:} All objects from the large container are now in the colored container. \\
        \midrule

        \textit{Insert In Peg} &
        \textit{Task Description:} Insert the square ring into the colored peg. \newline
        \textit{Success Criteria:} The square ring is in the colored peg. \\
        \midrule

        \textit{Meat Off Grill} &
        \textit{Task Description:} Pick the meat (chicken or steak) from the grill and place it into the designated area. \newline
        \textit{Success Criteria:} Meat is on the designated area. \\
        \midrule

        \textit{Open Bottle} &
        \textit{Task Description:} Remove the cap of the wine bottle. \newline
        \textit{Success Criteria:} Cap of the wine bottle is removed. \\
        \midrule

        \textit{Put Block} &
        \textit{Task Description:} Put the block in the target area. \newline
        \textit{Success Criteria:} The block is in the target area. \\
        \midrule

        \textit{Rubbish In Bin} &
        \textit{Task Description:} Put the rubbish in the bin. \newline
        \textit{Success Criteria:} Rubbish is in the bin. \\
        \midrule

        \textit{Stack Blocks} &
        \textit{Task Description:} Stack a specified number of colored blocks on the target block. \newline
        \textit{Success Criteria:} Specified number of blocks are stacked on top of the target block. \\
        \bottomrule
    \end{tabular}
\end{table*}
\subsection{Obtaining Visual Observations }
\label{app:vis_state}
\begin{figure*}[ht!]
  \centering
  \includegraphics[trim={0cm 8cm 0cm 0cm},clip, width=0.8\linewidth]{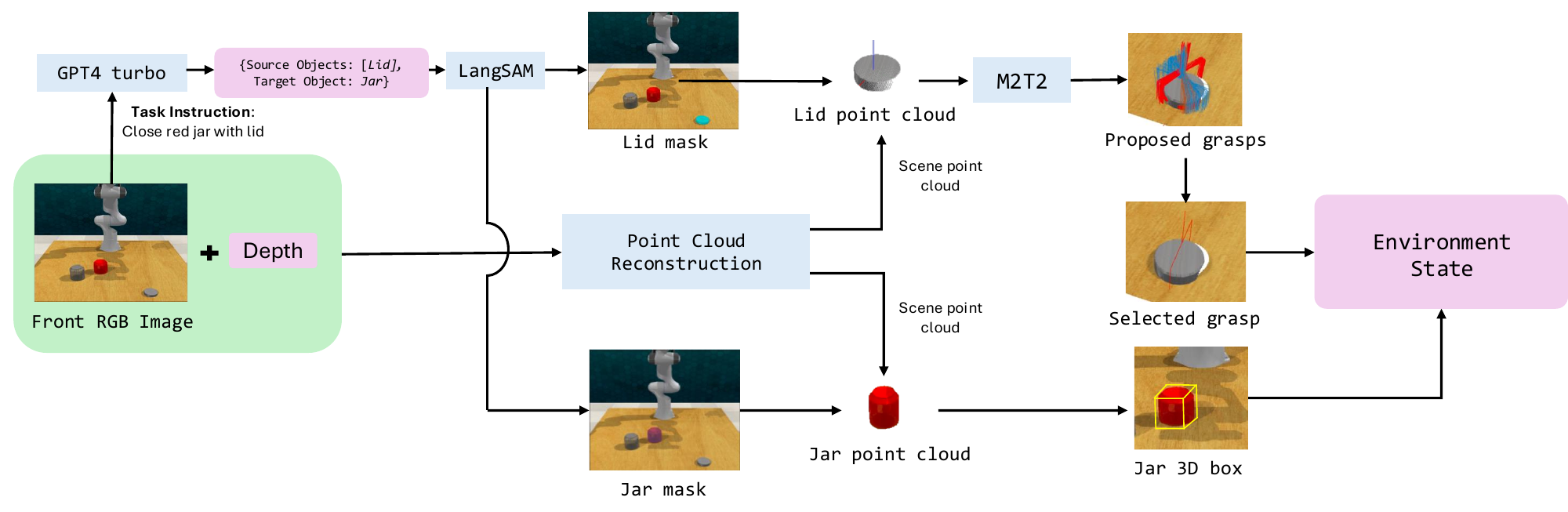}
  \caption{Process of obtaining visual observation, illustrated using the \textit{close jar} task. }
  \label{fig:VISUAL_STATES}
\end{figure*}

To obtain the visual observations, we use RGBD frames from the front view. First, we apply \textit{gpt-4-turbo}~\cite{openai2023gpt4} on the RGB image and task instruction, prompting it to generate appropriate names for the source and target objects, which is helpful for LangSAM to segment. Second, we generate 3D point clouds from the input RGBD frames. Then, we segment the 3D point clouds of the source and target objects by projecting the 2D masks generated by LangSAM~\cite{medeiros2024langsegmentanything} into 3D. Next, we employ the M2T2~\cite{Yuan2023M2T2MM} model to identify potential grasp poses on the source objects. We filter these grasp poses to retain only those with a top-down gripper orientation. Finally, we apply DBSCAN~\cite{ester1996density} clustering to the numerous grasp poses and select the central pose. For the placement, we use the 3D bounding box of the target object to identify the placement area. Thus, the selected grasp poses of the objects to be moved, along with the 3D bounding box of the target object, are used as the environment state. Figure~\ref{fig:VISUAL_STATES} illustrates the process of selecting the environment state, as described for the \textit{close jar} task with the instruction \textit{close red jar with lid}. In addition, to dynamically update the grasp and placement states, we introduce a history-aware mechanism to mitigate jitter in the extracted visual states (i.e., grasp and placement poses). Specifically, during each execution, if the translation distance is less than 0.01 meters or the change in gripper orientation is below 30 degrees, we retain the visual state from the previous step. This approach helps maintain stability during manipulation.

\clearpage 

\foreach \i in {1,...,35} { 
    \includepdf[pages={\i}, scale=0.7,
        pagecommand={
            \ifnum\i=1 
                \subsection{Execution Logs}\label{app:output}
               This section presents an example execution of our MALMM framework for the \textit{empty container} task. The initial setup is shown after the \textit{Task Instruction}, followed by images of changes captured after successful code execution.
                \vspace{1em} 
            \fi
            \vspace*{\fill} 
            \centering \captionof{figure}{Execution logs for MALMM method for the task \textit{empty container}.} 
            \vspace*{-0.5cm} 
        }
    ]{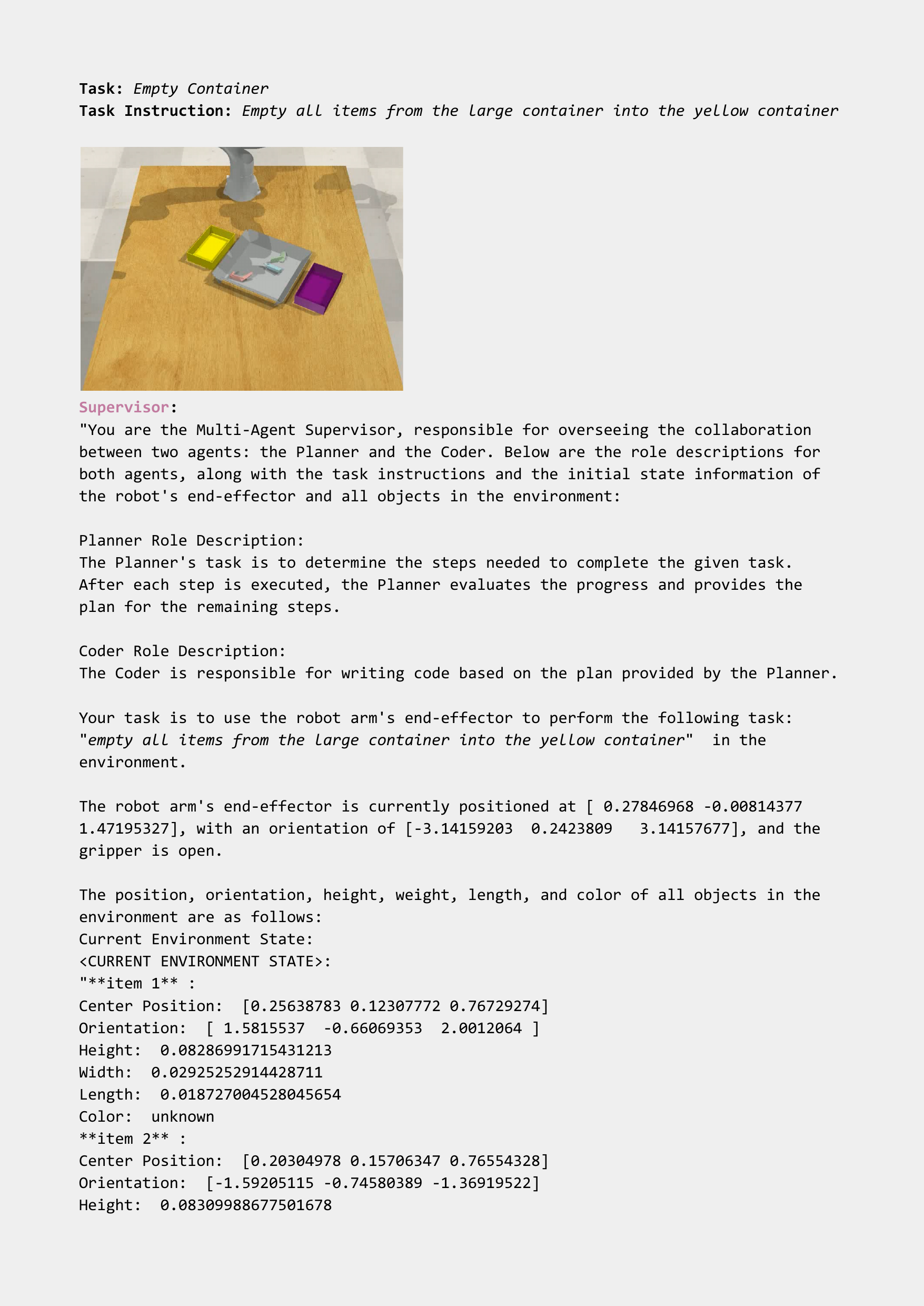}
}

\end{document}